\documentclass{article}

\newif\ifarxiv \arxivtrue
\newif\ifshowcomments \showcommentstrue

\usepackage{microtype}
\usepackage{graphicx}
\usepackage{subcaption}
\usepackage{float}
\usepackage{longtable}\usepackage{booktabs}\usepackage{tabularx,multirow}\providecommand{\jev}{Jev}

\usepackage{hyperref}

\usepackage{algorithm}
\usepackage{algorithmic}

\ifarxiv\PassOptionsToPackage{preprint}{icml2026}\fi
\usepackage{icml2026}
\ifarxiv\hypersetup{pdftitle={Retriever-Free Retrieval: Navigating Documents with Decision Models}, pdfsubject={}}\fi

\usepackage{amsmath}
\usepackage{amssymb}
\usepackage{mathtools}
\usepackage{amsthm}

\usepackage[capitalize,noabbrev]{cleveref}

\theoremstyle{plain}

\theoremstyle{definition}

\theoremstyle{remark}

\ifshowcomments\usepackage[textsize=tiny]{todonotes}\else\usepackage[disable,textsize=tiny]{todonotes}\fi

\icmltitlerunning{Retriever-Free Retrieval: Navigating Documents with Decision Models}

\newcommand{\TODO}[1]{\textcolor{red}{[TODO: #1]}}
\newcommand{\cw}[1]{\textcolor{teal}{Chao: #1}} 
\newcommand{\pj}[1]{\textcolor{blue}{Paul: #1}}
\newcommand{\method}{DecisionRAG}
\newcommand{\bl}[1]{\textcolor{olive}{#1}} 
\newcommand{\bd}[1]{\textcolor{violet}{#1}} \newcommand{\bdcomment}[1]{\bd{[Beixi: #1]}}
\ifshowcomments\else
  \renewcommand{\TODO}[1]{}\renewcommand{\cw}[1]{}\renewcommand{\pj}[1]{}\renewcommand{\bl}[1]{}
  \renewcommand{\bdcomment}[1]{}\renewcommand{\bd}[1]{#1}
\fi

\ifshowcomments\else\fi

\begin{document}

\twocolumn[
  \icmltitle{Retriever-Free Retrieval: Navigating Documents with Decision Models}

  \icmlsetsymbol{equal}{*}

  \begin{icmlauthorlist}
    \icmlauthor{Chao Wan}{equal,cornell}
    \icmlauthor{Paul Jünger}{equal,cornell}
    \icmlauthor{Beixi Du}{cornell}
    \icmlauthor{Bodong Liu}{cornell}
    \icmlauthor{Navaneetha Krishnan}{asapp}
    \icmlauthor{Kilian Q. Weinberger}{cornell,asapp}
  \end{icmlauthorlist}

  \icmlaffiliation{cornell}{Department of Computer Science, Cornell University, New York, USA}
  \icmlaffiliation{asapp}{ASAPP, New York, USA}

  \icmlcorrespondingauthor{Chao Wan}{cw862@cornell.edu}
  \icmlcorrespondingauthor{Paul Jünger}{pj287@cornell.edu}

  \icmlkeywords{Retrieval, Retrieval-Augmented Generation, Decision Models, Hierarchical Search}

  \vskip 0.3in
]

\printAffiliationsAndNotice{\icmlEqualContribution}

\begin{abstract}
While language models have become dramatically more capable, retrieval still largely relies on lexical matching or similarity in fixed embedding spaces. This creates a capability asymmetry: what the retriever fails to find, even the strongest generator cannot recover. We propose \method{}, which brings the capabilities of pretrained language models to retrieval using decision models that jointly evaluate candidate branches in a single forward pass, without generating any tokens, and show that judging candidates in a shared context makes their relevance estimates comparable. Like a person browsing a library, \method{} navigates the corpus’s inherent structure from broad topics down to individual passages, deciding at each level which branches are worth exploring. Without task-specific training or a first-stage retriever, \method{} outperforms dense retrieval with an 8B reranker by 13.8 R@1 points on MultiDoc2Dial and 17.4 Hit@1 points on Natural Questions. Across four benchmarks, including a corpus of 2.68 million Wikipedia paragraphs, retrieval takes just 0.7–1.8 seconds per query. 
\end{abstract}

\section{Introduction}

\suppressfloats[t]
\begin{figure}[t]
    \centering
    \includegraphics[width=\columnwidth]{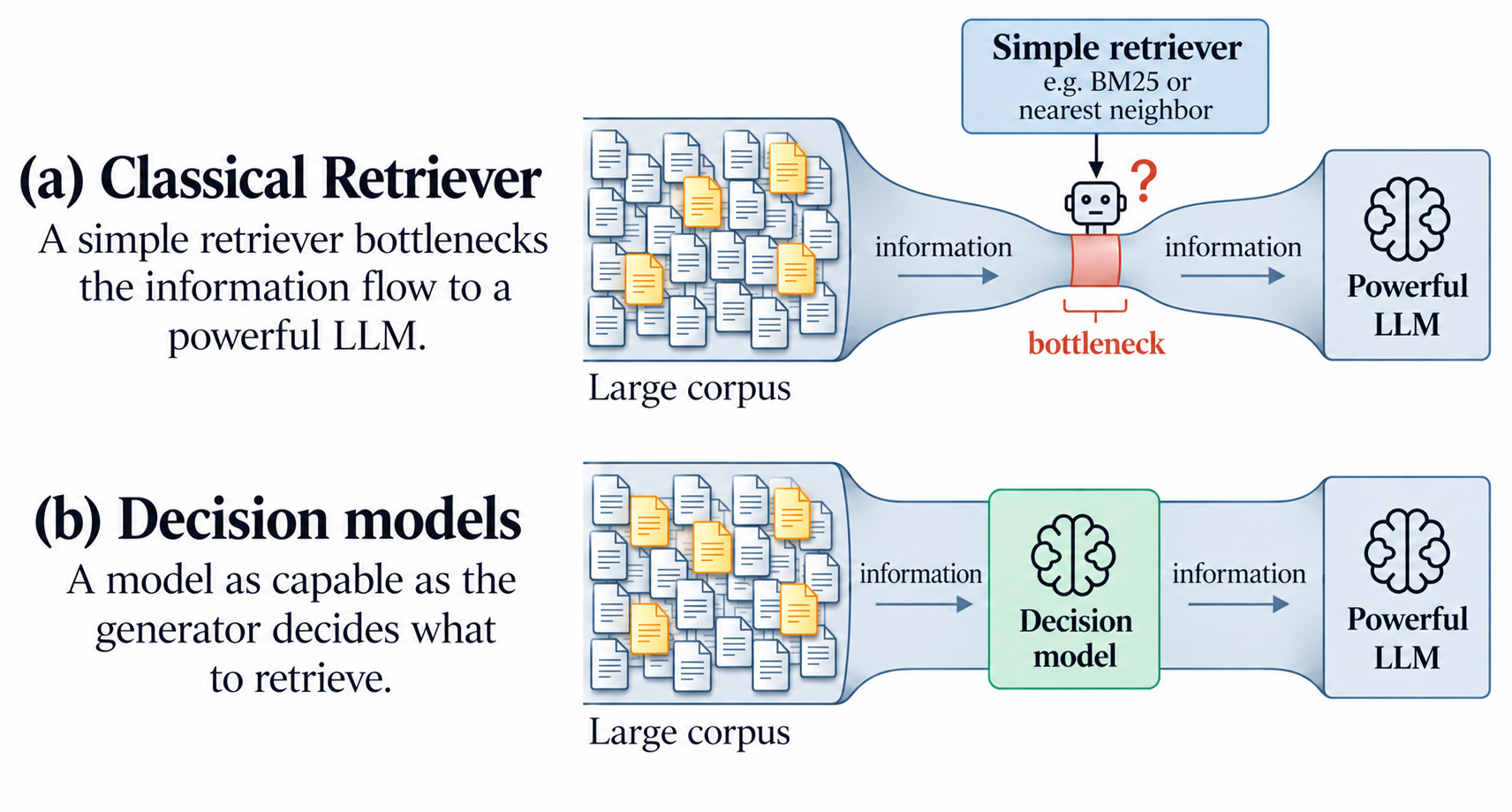}
    \caption{Capability asymmetry: a simple retriever bottlenecks a powerful LLM (a); decision models level the playing field (b).}
    \label{fig:motivation}
    \vspace{-14pt}
\end{figure}

Retrieval-augmented generation (RAG) has become a core component of modern language-model pipelines \citep{lewis2020rag, guu2020realm, gao2023retrieval}. By fetching evidence from an external corpus at inference time, RAG gives a model access to up-to-date information it never saw during training \citep{lewis2020rag, vu2024freshllms} and grounds its responses in verifiable sources \citep{shuster2021retrieval, gao2023enabling}.

Yet, while the language generation models have become dramatically more capable \citep{openai2026gpt5, anthropic2026opus}, the retrievers that decide what they see have not kept pace. Most retrieval systems still score each passage via heuristics based on term frequencies like BM25 \citep{robertson2009bm25}, or by nearest neighbors in a fixed embedding space~\citep{karpukhin2020dpr, izacard2022unsupervised, zhang2025qwen3embedding}. This creates a striking capability asymmetry between the retriever and the generation model (Figure~\ref{fig:motivation}): whatever the retriever misses, no amount of generator capability can recover \citep{barnett2024seven}.

Ideally, a model as capable as the generator would decide what to retrieve. Agentic approaches let an LLM navigate the corpus step by step~\citep{yao2023react, chen2023memwalker, jin2025searchr1}, but generating many tokens at every step in an agentic loop makes them too slow and costly for many applications \citep{jeong2024adaptiverag, yang2026searchagentx}. Rerankers offer a cheaper middle ground~\citep{nogueira2019passage, sun2023rankgpt, reddy2024first}, but they can only reorder what a first-stage retriever has already returned.

Decision models have emerged as a new paradigm to leverage the power of pretrained language models quickly without sampling any tokens \citep{typesafe2026jev, openai2026decisions, li2026llmasjev}.
Given a context and a set of questions, a decision model such as Jev \citep{typesafe2026jev} returns, for each question, a probability distribution over its answer options (for a yes/no question, $P(\mathrm{yes})$), answering many questions in parallel against a shared context in a single forward pass. This makes them a natural fit for retrieval, bringing the capability of LLMs to the choice of what to retrieve. Furthermore, scoring the candidates jointly makes their decisions comparable with one another \citep{chen2025icr, gupta2025blockrank}. Using decision models to choose what to retrieve thus promises to level the playing field between the retriever and generator.

\begin{figure*}[t]
    \centering
    \includegraphics[width=\textwidth]{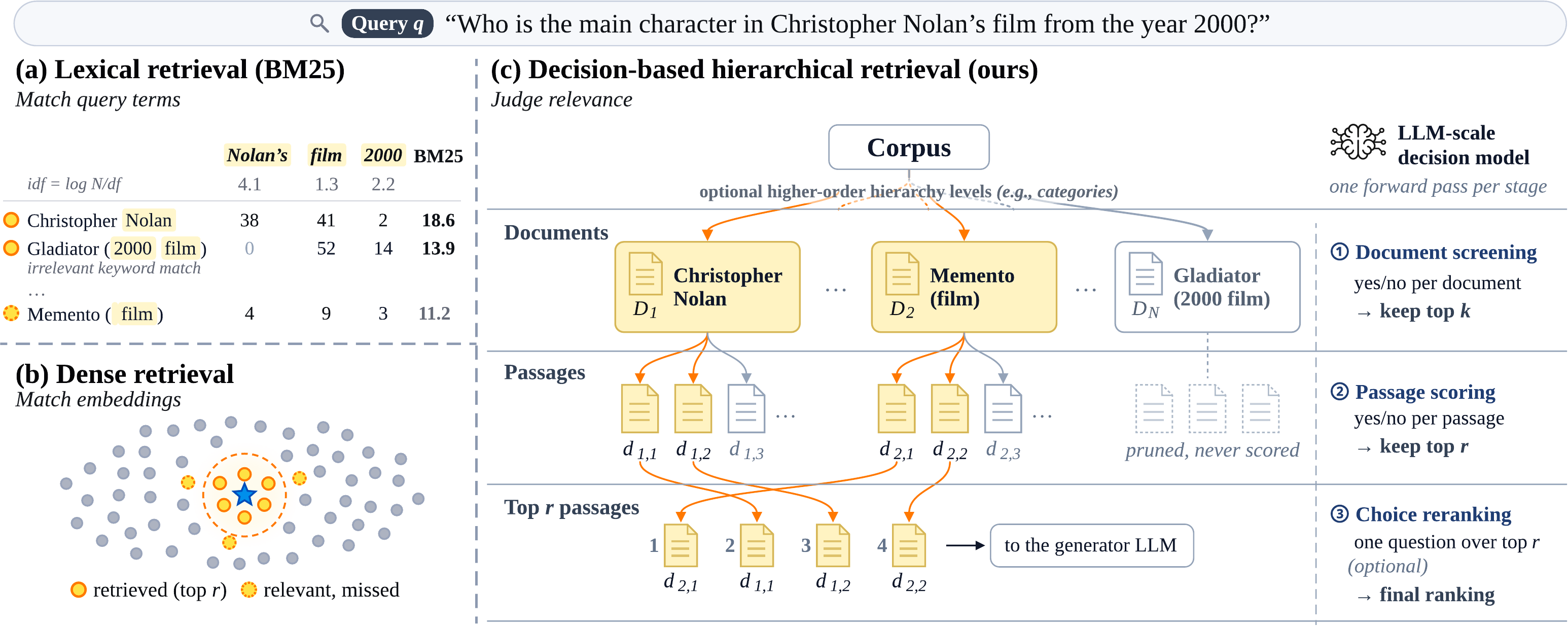}
    \caption{\textbf{Retrieval by matching versus retrieval by decisions.}
(a) BM25 and (b) dense retrieval match query terms or embeddings and can miss the relevant page.
(c) \method{} walks the corpus hierarchy: a decision model judges all candidates of a level in one forward pass, keeping the top $k$ documents and then the top $r$ passages; pruned documents are never expanded. Term frequencies, IDF values, BM25 scores, and rankings
in panel~(a) are illustrative.}
    \label{fig:method}
\end{figure*}

We propose \method{}, which retrieves by navigating instead of searching (Figure~\ref{fig:method}). Most corpora have an inherent structure of topics, documents, sections, and passages, which humans exploit to find information efficiently \citep{pirolli1999information}. \method{} leverages this structure through hierarchical decisions \citep{geva2018docqn, liu2021dense}: at each level, a decision model jointly evaluates candidates to identify the most promising branches, progressively narrowing the search from broad categories (e.g. topics or document titles) down to individual passages. Each decision requires only a single forward pass without generating any tokens, keeping retrieval fast even for large corpora. \method{} thus brings the capabilities of pretrained language models directly to retrieval, without a first-stage retriever at query time, a precomputed index, or task-specific training.

We evaluate \method{} on four benchmarks spanning multi-turn dialogue \citep{feng2021multidoc2dial, katsis2025mtrag}, scenario-based questions \citep{sun2022conditionalqa}, and open-domain search over 2.68 million Wikipedia paragraphs \citep{kwiatkowski2019natural, wang2024dapr}. Without task-specific training, \method{} outperforms dense retrieval with an 8B reranker by 13.8 R@1 points on MultiDoc2Dial and 17.4 Hit@1 points on Natural Questions. Judging candidates in a shared context contributes to these gains, improving accuracy by up to 7.7 points over independent judgments (\cref{sec:ablations:shared}). Despite harnessing the capabilities of a language model at every stage, \method{} retrieves in just 0.7–1.8 seconds at a cost of 0.2–2.3 cents per query. Compared with an LLM that generates navigation decisions over the same hierarchy token-by-token, it achieves comparable accuracy with 34–145$\times$ lower latency (\cref{sec:ablations:generative}). Training a dedicated 8B decision model on in-domain data further raises R@1 on MultiDoc2Dial to 60.7, surpassing the best compared published task-trained retriever by 5.7 points and suggesting substantial headroom as decision models improve.
\section{Related Work}
\label{sec:related}

Traditional retrieval methods, including BM25 \citep{robertson2009bm25}
and dense retrievers \citep{karpukhin2020dpr}, rank passages using
lexical matching or learned embedding similarity. Neural reranking
methods refine these rankings using language-model-based relevance
estimates. Cross-encoders evaluate individual query--passage pairs
\citep{nogueira2020monot5}, while LLM-based rerankers perform listwise
ranking by generating permutations \citep{sun2023rankgpt} or iteratively
selecting candidates from subsets
\citep{zhuang2024setwise,li2026wholepool}.
Although reranking can improve retrieval quality, it remains limited
by the candidate recall of the initial retrieval stage.

An alternative line of work integrates language models directly into
the retrieval process. Agentic approaches interleave reasoning with the use of
search-engines to iteratively identify relevant evidence
\citep{yao2023react,jin2025searchr1}. Other methods use hierarchical
representations to guide retrieval. MemWalker navigates a hierarchy
of summaries constructed from long documents \citep{chen2023memwalker},
while PageIndex uses document tables of contents for LLM-guided
navigation \citep{vectify2026pageindex}. LATTICE extends hierarchical
navigation to corpus-level retrieval, constructing a semantic tree
and using LLM-generated relevance assessments to select branches
at each level \citep{gupta2026lattice}.
These approaches require generating tokens sequentially at each
retrieval step, which can incur substantial latency when navigation
involves multiple rounds of reasoning.

Recent work has explored more efficient approaches to
language-model-based ranking. FIRST derives listwise rankings
from the logits of a single generated identifier
\citep{reddy2024first}, while ICR and BlockRank extract ranking
signals from attention patterns over jointly presented candidates
\citep{chen2025icr,gupta2025blockrank}.
Decision models offer another approach to evaluating multiple
candidates without generating tokens
\citep{typesafe2026jev,openai2026decisions}.
Given a shared context and a set of questions, these models produce
probability distributions over the answer options of each question
in a single forward pass. This enables parallel relevance judgments
over multiple candidates, with each candidate evaluated against
the same contextual information.

Hierarchical retrieval has also been studied as a means of reducing
the search space in large document collections. DocQN learns to
navigate document trees using reinforcement learning
\citep{geva2018docqn}, while dense hierarchical retrieval
progressively selects documents and passages using trained
encoders \citep{liu2021dense}. RAPTOR constructs a hierarchy of
recursively generated summaries to support retrieval at different
levels of abstraction \citep{sarthi2024raptor}.
Related work has also highlighted the importance of document-level
context for passage retrieval \citep{wang2024dapr,xu2025rdr2,sfre2g2026}.

\method{} combines hierarchical retrieval with non-generative relevance
judgments. It uses existing document structure where available and
constructs document groups otherwise (\cref{sec:setup:data}).
Unlike conventional reranking pipelines, \method{} does not require
a first-stage retriever to construct a candidate pool. Unlike LLM-guided hierarchical search, it evaluates candidates at each
level through parallel relevance judgments in a single forward pass,
without generating tokens.

\section{Method}
\label{sec:method}

\subsection{Problem Formulation}
We consider a corpus \(\mathcal D=\{
D_1,\ldots,D_N\}\), where each document \(D_i\) consists of passages \(d_{i1},\ldots,d_{iM_i}\). Given a query \(q\), which may include the preceding dialogue turns, our goal is to retrieve the passages containing the evidence needed to answer it. (\cref{fig:method}) More generally, we view the corpus as a tree whose leaves are passages and whose internal nodes group them at increasing granularity, such as help centers, topic clusters, documents, or section headings. Each node has a short textual description: the text itself for a passage, the title for a document, and a one-sentence summary for a cluster. We call each depth of the tree a \emph{level} and each round of requests to the decision model a \emph{stage}; most levels take one stage, and the final reranking is a stage of its own. As a running example, we use the query 
\begin{equation}
\label{eq:question}
\text{\parbox{0.85\linewidth}{\centering
\textit{    Who is the main character in Christopher Nolan's film from the year 2000?}
}}
\end{equation}
which can be answered by the Wikipedia corpus.

\subsection{Relevance Judgments with Decision Models}
\label{sec:method:judgments}
\label{sec:preliminaries}

\begin{figure}[H]
    \centering
    \begin{subfigure}[b]{\columnwidth}
        \centering
        \includegraphics[width=0.95\columnwidth]{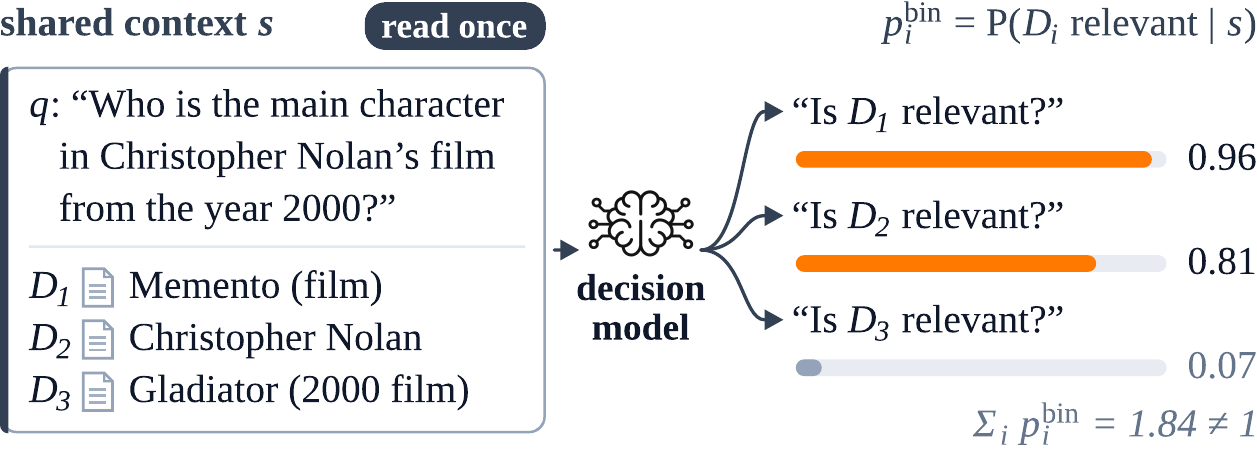}
        \caption{Parallel binary judgments for documents $D_i$ against a shared context $s_D$.}
        \label{fig:binary}
    \end{subfigure}\\[4pt]
    \begin{subfigure}[b]{\columnwidth}
        \centering
        \includegraphics[width=0.95\columnwidth]{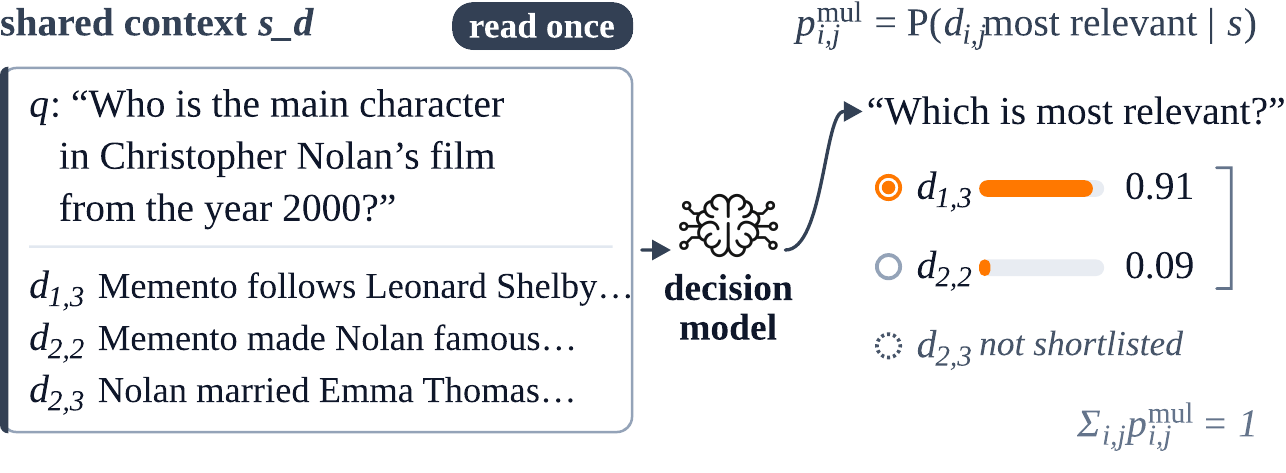}
        \caption{Multiple-choice reranking of the top passages from the binary judgments against the same shared context $s$; the probabilities sum to one.}
        \label{fig:choice}
    \end{subfigure}
    \caption{\textbf{The two judgment modes} of a decision model, each against a shared context that the model reads once.}
    \label{fig:judgments}
\end{figure}

Decision models such as \jev{} \citep{typesafe2026jev} and OpenAI's decision
model \citep{openai2026decisions} evaluate questions against a context and
return probabilities over predefined answer options, without autoregressive
generation. We use two of their inference modes, parallel binary judgments
and multiple-choice judgments, both against a shared context $s$
(\cref{fig:judgments}).

\paragraph{Parallel binary judgments.}
Given a shared context $s$ and a set of binary questions
$\mathcal Q=\{Q_1,\ldots,Q_n\}$, the decision model
evaluates all questions in parallel in a single forward
pass. For each question $Q_i$, it produces a probability
of answering ``Yes'',
\begin{equation}
\label{eq:binary}
p_i^{\mathrm{bin}} = P(\mathrm{Yes}\mid s,Q_i).
\end{equation}
Each question is evaluated against the same context,
but its answer probabilities are normalized only over
``Yes'' and ``No'', not across questions. Thus, the
resulting scores need not sum to one.
Binary judgments therefore impose no limit on the number of candidates
beyond the model's context length; larger levels are split across
parallel requests (\cref{sec:method:practice}).

\paragraph{Multiple-choice judgments.}
Given a context $s$, a question $Q$, and a set of
answer options $\mathcal A=\{a_1,\ldots,a_m\}$,
the decision model produces a probability distribution
over the options,
\begin{equation}
\label{eq:choice}
p_i^{\mathrm{mul}}
= P(a_i\mid s,Q,\mathcal A),
\qquad
\sum_{i=1}^{m}p_i^{\mathrm{mul}}=1.
\end{equation}
Unlike parallel binary judgments, multiple-choice
judgments evaluate the answer options as competing
alternatives within a single decision.
The number of answer options is limited (255 for \jev{}), so we apply
this mode only to the top candidates from the binary judgments.

\paragraph{Judging candidates in a shared context.}
Because every question sees the same context, each
probability is formed in view of the alternatives: a candidate
that would look relevant on its own receives a lower
$P(\mathrm{Yes})$ when a clearly better one is present. The
shared context thus puts the individual probabilities into
perspective, making them comparable across candidates.
Our ablations support this: scoring candidates together rather than
separately improves R@1 on MultiDoc2Dial by 7.7 points and Hit@1 on Natural
Questions by 6.5 points (\cref{sec:ablations:shared,sec:ablation:ctxsize}).
They also show where the comparison matters: splitting a pool into smaller
requests lowers Hit@1 but leaves R@10 unchanged. Scores from separate requests
therefore suffice for screening, where recall counts, whereas the final
ranking benefits from one shared request.

\subsection{Hierarchical Retrieval}
\label{sec:method:navigation}

\begin{algorithm}[t]
\caption{\method{}}
\label{alg:nav}
\begin{algorithmic}[1]
\REQUIRE query $q$; corpus tree with $L$ levels; budgets $k_1,\ldots,k_{L-1}$; number of returned passages $r$; reranking shortlist size $m \le r$
\STATE $\mathcal C \leftarrow$ children of the root
\FOR{$\ell = 1,\ldots,L-1$}
  \STATE $p^{\mathrm{bin}} \leftarrow \textsc{BinaryJudge}(q,\mathcal C)$ \COMMENT{\cref{eq:binary}}
  \STATE $\mathcal C \leftarrow \bigcup_{c\in\text{top-}k_\ell(\mathcal C,\,p^{\mathrm{bin}})} \text{children}(c)$
\ENDFOR
\STATE $p^{\mathrm{bin}} \leftarrow \textsc{BinaryJudge}(q,\mathcal C)$
\STATE $R \leftarrow \text{top-}r(\mathcal C,\,p^{\mathrm{bin}})$
\IF{multiple-choice reranking is enabled}
  \STATE $p^{\mathrm{mul}} \leftarrow \textsc{ChoiceJudge}(q,R[1{:}m])$ \COMMENT{\cref{eq:choice}}
  \STATE reorder $R[1{:}m]$ by decreasing $p^{\mathrm{mul}}$
\ENDIF
\STATE \textbf{return} $R$
\end{algorithmic}
\end{algorithm}

Rather than scoring every passage in the corpus individually, \method{} progressively narrows the
search space by traversing a corpus hierarchy.
At each level, a decision model evaluates the
candidates under consideration and retains the
most promising ones for further exploration.
Candidates not selected are pruned, and their
descendants are never evaluated. 

We illustrate this process in \cref{fig:method} (c) using the running example in \cref{eq:question}. 

\paragraph{Document screening.}
We first evaluate candidate documents using parallel binary
judgments (\cref{eq:binary}). The shared context $s_D$
contains the query and descriptions of all candidate
documents, while each binary question asks whether a
particular document is worth exploring for the query.
As illustrated in \cref{fig:binary}, the model evaluates all candidates in a single forward pass, assigning each a relevance score. In our running example, the pages for \textit{Christopher
Nolan} and \textit{Memento (film)} are retained,
while \textit{Gladiator (2000 film)} is discarded despite the latter's keyword overlaps with the query. 
We retain the top $k$ documents according to their
relevance scores and prune the remaining branches.

\paragraph{Passage scoring.}
We next expand the retained documents into their
passages and evaluate them using parallel binary judgments
(\cref{eq:binary}). The shared context $s_P$ now contains
the query and all candidate passages, allowing the model
to assess their relevance in view of one another, so that
their probabilities are directly comparable
(\cref{sec:method:judgments}).
We rank the passages by their binary relevance scores
and keep the top $r$ across all selected documents. On Natural Questions, where the paragraph stage spans 30 pages, each paragraph's score is first multiplied by its page's  (\cref{app:retrieval:nq}).
In our running example, a passage from
\textit{Memento (film)} describing its protagonist,
Leonard Shelby, provides the evidence needed to answer
the query.

\paragraph{Optional passage reranking.}
To refine the final ranking, we optionally apply a multiple-choice
judgment over the top $m \le r$ passages selected by binary scoring (\cref{eq:choice}). The choice context lists only these $m$ passages
(\cref{app:retrieval:choice}). Their choice 
probabilities determine their order, while the remaining passages
retain their binary ranking. We use $m=20$, except on Natural
Questions, where $m=5$. On MTRAG, we instead promote only the
highest-probability choice to rank one, preserving the relative
order of all other passages.
As illustrated in \cref{fig:choice}, the model compares the
shortlisted passages as competing alternatives within a single
decision.

Binary judgments suit the inner levels: they have no option limit, run in
parallel, and favor recall, which matters while the search narrows. The
multiple-choice judgment is reserved for the end, where it compares the top
passages from the binary judgments directly and sharpens the top rank
(\cref{sec:ablations:choice}). \Cref{alg:nav} summarizes the procedure for a
hierarchy of any depth.

\paragraph{Budgets and pruning.}
Pruning is irreversible: a branch dropped at one level is not revisited for
this query. The budgets $k_\ell$ therefore lean towards recall, trading
coverage for cost (\cref{sec:ablations:budget}). A budget can also be a
probability threshold, e.g., keeping a runner-up help center whenever the top
score is below 0.8 (\cref{app:retrieval:md2d}).

\subsection{Hierarchy Construction and Scaling}
\label{sec:method:practice}

\paragraph{Higher-order hierarchies.}
The document--passage hierarchy is only the simplest
instance of \method{}. For larger corpora, documents
can be organized into higher-order groups, such as
topics, categories, or products, using existing corpus
structure or grouping related documents.
The same procedure then applies recursively: at each
level, parallel binary judgments identify promising
groups, and only their children are evaluated at the
next level. For example, when searching the full
Wikipedia corpus, one could traverse categories
related to films and Christopher Nolan before reaching
the individual page for \textit{Memento}.
Thus, \method{} extends to deeper hierarchies without
requiring all documents to be evaluated at once.
When a corpus lacks such structure, a hierarchy can be
built offline by clustering document embeddings, e.g.,
with recursive $k$-means \citep{nister2006vocabtree} or
scalable agglomerative clustering
\citep{bateni2017affinity,monath2021scc,dhulipala2023terahac},
as we do for Natural Questions
(\cref{sec:setup:data}). The tree need not be uniform: on MTRAG, a heading
level is inserted only below pages with more than 60 passages.

\paragraph{Levels too large for one request.}
A level may hold more candidates than fit into one request; on Natural
Questions, the 150 retained clusters contain about 17,500 page titles. We then
split the level into parallel requests, each with its own shared context. This
costs little: the benefit of the shared context levels off at about 50
candidates per request (\cref{sec:ablation:ctxsize}), and pooled scores
suffice for screening. Where the top candidates of a split level should be
compared more closely, we rescore them in one request, as for the top 50 pages
on MTRAG (\cref{app:retrieval:mtrag}).

\paragraph{Combining with a retriever.}
\method{} needs no first-stage retriever, but it can use one: adding the top
passages of a dense retriever to the final candidates recovers evidence that
pruning discarded (\emph{+ dense candidates}; \cref{sec:results:hybrid}).

\begin{figure}[H]
    \centering
    \includegraphics[width=\columnwidth]{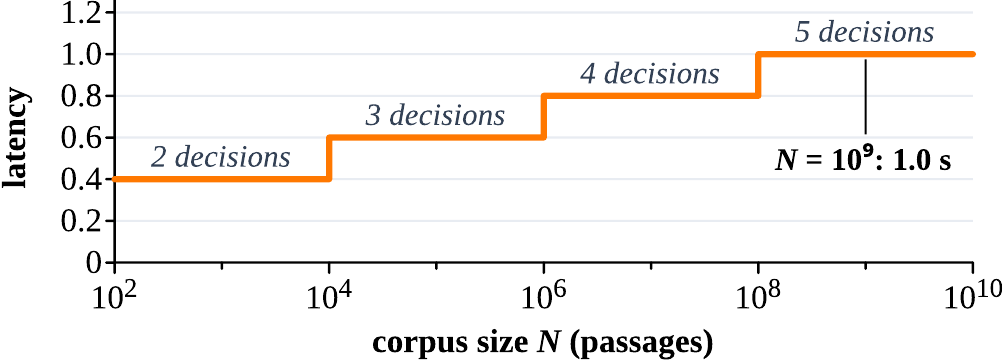}
    \caption{\textbf{Analytical model: sequential stages grow with the depth of the hierarchy, not the size of the corpus.}
     Theoretic latency per query (\cref{eq:scaling}, $b=100$), assuming a balanced hierarchy and enough serving capacity to judge every level in parallel: each $100\times$ more passages adds one level, i.e., one more sequential decision (${\approx}0.2$\,s).}
    \label{fig:scaling}
\end{figure}

\paragraph{Scaling with corpus size.}
Because all candidates of a level are judged in parallel,
latency depends on how many levels the hierarchy has, not on
how many passages the corpus holds. Assume a roughly balanced
hierarchy in which every node groups at most $b$ children, e.g.,
topics of 100 documents with 100 passages each. Every additional
level then multiplies the number of reachable passages by up to $b$,
so a corpus of $N$ passages needs about $\lceil\log_b N\rceil$
levels. Each level adds one sequential
decision, so the latency per query is
\begin{equation}
T(N)\approx t_{\mathrm{req}}\lceil\log_b N\rceil,
\label{eq:scaling}
\end{equation}
where $t_{\mathrm{req}}\approx0.2$\,s is the latency of one
request. With $b=100$, even $10^9$ passages need only five
decisions, about one second (\cref{fig:scaling}). Cost, in contrast, grows with
the number of judgments: $b$ children for each of the $k$
nodes kept per level. \Cref{eq:scaling} bounds the sequential depth rather
than the total compute, but because each level keeps only its
top $k$ nodes, no level scores more than $k \cdot b$ candidates.

 \section{Experimental Setup}
\label{sec:setup}

We evaluate \method{} on four benchmarks, ranging from help-center dialogues over 413 pages to open-domain questions over 108,626 Wikipedia pages (\cref{tab:datasets}). Unless stated otherwise, we use the pretrained \jev{} decision model (TypeSafe’s Jev 1.13, \texttt{typesafe/jev-1.13}) \citep{typesafe2026jev}, without task-specific fine-tuning or a conventional first-stage retriever. We also evaluate task-trained, open-weight decision models in \cref{sec:results:trained}.

\begin{table*}[t]
\centering
\small
\caption{Overview of the four retrieval benchmarks: MultiDoc2Dial
\citep{feng2021multidoc2dial}, ConditionalQA \citep{sun2022conditionalqa},
MTRAG \citep{katsis2025mtrag} and Natural Questions
\citep{kwiatkowski2019natural} in its DAPR version \citep{wang2024dapr}.}
\label{tab:datasets}
\begin{tabular}{@{}llll@{}}
\toprule
Dataset & Corpus hierarchy & Evaluation set & Gold evidence \\
\midrule
MultiDoc2Dial
  & 4 centers / 413 pages / 4,110 passages
  & 4,201 turns
  & 1 passage \\

ConditionalQA
  & 652 pages / 58,483 paragraphs
  & 285 questions (271 labeled)
  & 1--22 paragraphs \\

MTRAG (IBM Cloud)
  & 172 products / 8,578 pages / 72,442 passages
  & 205 turns (188 labeled)
  & 1--7 passages \\

Natural Questions (DAPR)
  & 1,000 clusters / 108,626 pages / 2.68M paragraphs
  & 500 test questions
  & 1--3 paragraphs \\
\bottomrule
\end{tabular}
\end{table*}

\subsection{Datasets and Corpus Construction}
\label{sec:setup:data}

\textbf{MultiDoc2Dial} \citep{feng2021multidoc2dial}
contains goal-oriented dialogues grounded in US government
documentation. \textbf{ConditionalQA}
\citep{sun2022conditionalqa} contains scenario-based questions
about UK government guidance, often requiring evidence from
multiple paragraphs. \textbf{MTRAG}
\citep{katsis2025mtrag} evaluates conversational retrieval
over technical documentation; we use its IBM Cloud subset.
\textbf{Natural Questions} \citep{kwiatkowski2019natural}
contains real search queries grounded in Wikipedia.

For the first three datasets, we construct corpus hierarchies
from the existing document structure and metadata. For Natural Questions,
whose Wikipedia corpus lacks a hierarchy above the page,
we cluster pages into 1,000 groups using $k$-means over
e5-base-v2 embeddings \citep{wang2022text} of page titles
and first sentences. Each cluster is described by a
one-sentence summary generated by Qwen3-8B
\citep{yang2025qwen3technical} from its page titles.

\subsection{Retrieval Configurations}
\label{sec:setup:walk}

We select dataset-specific navigation hierarchies and retention budgets using separate design sets (\cref{app:datasets:design}). For conversational benchmarks, we include the last six dialogue turns and maintain the notepad described in \cref{app:retrieval}.

On \textbf{MultiDoc2Dial}, we navigate a
\emph{help center $\rightarrow$ page $\rightarrow$ passage}
hierarchy, retaining the top ten pages and scoring their
passages alongside those from up to two notepad pages.

On \textbf{ConditionalQA}, we use a
\emph{page $\rightarrow$ paragraph} hierarchy,
retaining the top three pages and jointly scoring their
paragraphs.

On \textbf{MTRAG}, we use a
\emph{page $\rightarrow$ heading $\rightarrow$ passage}
hierarchy, scoring all page titles in parallel, jointly rescoring
the top 50 pages, and retaining the top ten. For long pages,
we select relevant headings before scoring their passages.

On \textbf{Natural Questions}, we navigate a
\emph{cluster $\rightarrow$ page $\rightarrow$ paragraph}
hierarchy, retaining 150 of the 1,000 clusters, evaluating paragraphs
from the top 30 pages.

All \method{} (\jev{}) configurations finish with multiple-choice ranking of the top 20 passages (top five for Natural Questions); on MTRAG, only the selected passage is promoted to rank one. The \emph{+ dense candidates} variant adds the dense top 20 passages (top ten on ConditionalQA) to the final candidate pool. Additional routing rules and implementation details are provided in \cref{app:retrieval}.

We also evaluate task-trained decision models based on Qwen3-Base at 1.7B, 4B, and 8B parameters (\cref{sec:results:trained}). Architecture and training details are provided in \cref{sec:setup:train}.

\subsection{Baselines}
\label{sec:setup:baselines}

We compare \method{} with retrieval methods used without
task-specific fine-tuning, as well as task-trained methods
where available. The baselines are grouped as follows:

\paragraph{First-stage retrieval.}
We compare against BM25 with stemming and stopword removal,
SPLADE++ \citep{formal2022spladepp}, and
Qwen3-Embedding-8B \citep{zhang2025qwen3embedding}.
On ConditionalQA and MTRAG, the appendix also reports
SPLADE-v3 with a DistilBERT encoder \citep{lassance2024spladev3}
(\cref{tab:cqa-extended,tab:mtrag-judged}).

\paragraph{Reranking.}
We evaluate bge-reranker-v2-m3
\citep{chen2024bgem3,baai2024rerankerv2m3} and
Qwen3-Reranker-8B on the top 100 passages retrieved by
Qwen3-Embedding-8B. We also evaluate GPT-5
\citep{openai2026gpt5} with low reasoning effort,
reranking the top 20 dense candidates using a
RankGPT-style prompt \citep{sun2023rankgpt}.

\paragraph{Published reference results.}
We include the MultiDoc2Dial authors’ fine-tuned DPR \citep{karpukhin2020dpr,feng2021multidoc2dial} and DPR with cross-encoder \citep{sfre2g2026}, as well as the ELSER retrieval results released with MTRAG \citep{katsis2025mtrag}. These results are included as references rather than rerun under our evaluation setup.

\begin{table*}[t]
\centering
\small
\setlength{\tabcolsep}{2.5pt}

\caption{Retrieval without task-specific training on four benchmarks (\%).}

\label{tab:retrieval-summary}
\begin{tabular*}{\textwidth}{@{\extracolsep{\fill}}lrrrrrrrrr@{}}
\toprule
& \multicolumn{2}{c}{MultiDoc2Dial} & \multicolumn{2}{c}{ConditionalQA} & \multicolumn{2}{c}{MTRAG$^\ddagger$} & \multicolumn{3}{c}{Natural Questions} \\
\cmidrule(lr){2-3}\cmidrule(lr){4-5}\cmidrule(lr){6-7}\cmidrule(lr){8-10}
Method & R@1 & R@10 & Hit@1 & R@10 & R@5 & nDCG@5 & Hit@1 & R@10 & nDCG@10 \\
\midrule
\multicolumn{10}{@{}l}{\textit{Conventional retrieval pipelines}} \\
BM25 & 20.6 & 56.4 & 24.7 & 35.0 & 18.8 & 30.9 & 16.4 & 42.3 & 28.4 \\
SPLADE++ & 24.4 & 58.8 & 28.4 & 38.3 & 27.8 & 47.1 & 36.0 & 74.6 & 54.5 \\
Dense & 27.9 & 70.9 & 47.2 & 54.0 & 26.0 & 46.5 & 46.2 & 87.2 & 66.0 \\
Dense top-100 + BGE & 22.8 & 63.9 & 53.1 & 60.6 & 14.5 & 25.3 & 54.8 & 88.7 & 71.7 \\
Dense top-100 + Qwen3-Reranker & 33.6 & 81.1 & 64.6 & 67.2 & 32.5 & 56.5 & 48.0 & 89.0 & 69.0 \\
\midrule
\multicolumn{10}{@{}l}{\textit{LLM-based listwise reranking}} \\
Dense top-20 + GPT-5 & 47.5 & 78.6 & 73.4 & 64.0 & 31.8 & 59.4 & 60.8 & 88.9 & 74.6 \\
\midrule
\multicolumn{10}{@{}l}{\textit{Decision-based retrieval (ours)}} \\
\textbf{\method{} (\jev{})} & 47.4 & 83.6 & 69.7 & 71.9 & 34.2 & 62.3 & 65.4 & 86.7 & 76.0 \\
\textbf{\method{} (\jev{}) + dense candidates} & 48.5 & 84.7 & 73.4 & 76.3 & 36.6 & 65.0 & 66.4 & 88.9 & 77.6 \\
\bottomrule
\end{tabular*}

\vspace{2pt}
\begin{minipage}{\textwidth}
\footnotesize
Dense: Qwen3-Embedding-8B; BGE: bge-reranker-v2-m3.
Rerankers rerank the dense top 100, GPT-5 the dense top 20.
\emph{+ dense candidates} adds the
dense top 20 (top 10 on ConditionalQA) to its last stage.
Natural Questions: all 2.68M paragraphs are searched.
$^\ddagger$MTRAG: gold and GPT-5-judged relevant passages, identical texts counted once (\cref{app:datasets:mtrag}); conventional baselines use their best input, the last turn (BGE: six turns).
\end{minipage}
\end{table*}

\subsection{Evaluation Metrics and Protocol}
\label{sec:setup:metrics}

We denote Recall@k as R@k and report conventional retrieval metrics on corresponding benchmarks in \cref{tab:retrieval-summary}.
Additional metrics are reported in the appendix (\cref{app:full-results}). 

To assess downstream question answering, we use 
Qwen3.5-27B to answer each query from a system's top five
passages. We report exact match on ConditionalQA and
Natural Questions and LLM-judged answer acceptance on MTRAG, whose references are long
replies
(\cref{app:rag}).

Confidence intervals are paired bootstrap 95\%
intervals (10,000 resamples of whole dialogues,
conversations or questions). We also measure per-query
cost and online latency (\cref{app:timing}).

\section{Results}
\label{sec:results}

We evaluate \method{} across four benchmarks, comparing its retrieval quality against conventional retrieval and reranking pipelines. We further assess downstream answer quality, the effects of task-specific training, and online latency and cost.

\subsection{Retrieval Effectiveness}
\label{sec:results:main}

Without task-specific training or a conventional retrieval index, \method{} substantially outperforms dense retrieval with an 8B reranker (\cref{tab:retrieval-summary}): On MultiDoc2Dial, it improves R@1 by 13.8 percentage points, and on Natural Questions, it improves Hit@1 by 17.4 points.

Natural Questions also shows that \method{} needs no ready-made hierarchy:
Wikipedia's 108,626 pages, far too many for one top level, are simply clustered
into 1,000 groups (\cref{sec:setup:data}). \method{} then reads only 16\% of
the page titles. Any corpus can
be given a hierarchy this way, with a number of sequential stages that grows only logarithmically with its
size (\cref{eq:scaling,sec:ablations:budget}).

Even when using a strong language model such as GPT-5 to rerank the dense retrieval results, \method{} achieves comparable or higher rank-one accuracy on MultiDoc2Dial and Natural Questions (65.4 versus 60.8). It also improves ConditionalQA R@10 (71.9 versus 64.0) and MTRAG R@5 (34.2 versus 31.8), although GPT-5 achieves higher ConditionalQA Hit@1 (73.4 versus 69.7).
On MTRAG, whose annotated evidence was collected with ELSER, we judge the top five passages of every system, and \method{} ranks first among all eighteen, including last-turn and rewritten-query baselines (R@5 34.2 versus 32.5 for the best baseline; \cref{app:datasets:mtrag,tab:mtrag-judged}).

\label{sec:results:hybrid}
\method{} can also complement dense retrieval by incorporating
dense-retrieved passages into its final ranking stage.
This combination improves R@10 from 83.6 to 84.7 on
MultiDoc2Dial, from 71.9 to 76.3 on ConditionalQA,
and from 86.7 to 88.9 on Natural Questions.
These gains suggest that the two approaches recover
complementary evidence and can be effectively combined.

\subsection{Retrieval-Augmented Generation}
\label{sec:results:rag}

\begin{table}[h]
\centering\small
\setlength{\tabcolsep}{3pt}
\caption{Answer quality of one generator (Qwen3.5-27B) given each system's top
five passages (\%): exact match, and on MTRAG the share of replies judged
correct (188 labeled turns). Qwen3-RR: Qwen3-Reranker-8B; NQ: Natural Questions.}
\label{tab:rag}
\begin{tabular}{@{}lrrr@{}}
\toprule
& ConditionalQA & MTRAG & NQ \\
Retrieval & EM & Judged & EM \\
\midrule
None & 1.8 & 6.4 & 26.9 \\
Dense & 43.5 & 31.4 & 53.0 \\
Dense top-100 + Qwen3-RR & 45.4 & 39.4 & 52.4 \\
Dense top-20 + GPT-5 & 44.3 & 34.0 & 54.5 \\
\textbf{\method{} (\jev{})} & 46.9 & 44.1 & 54.2 \\
\quad \textbf{+ dense candidates} & 47.6 & 47.3 & 54.8 \\
\midrule
Gold passages & 49.4 & 60.1 & 65.3 \\
\bottomrule
\end{tabular}
\end{table}

We next ask whether the retrieved evidence improves downstream answers. We provide each method’s top five passages to the same generator model, Qwen3.5-27B (\cref{sec:setup:metrics,tab:rag}). The largest improvement occurs on MTRAG, where \method{} increases judged answer correctness from 39.4\% with Qwen3-Reranker to 44.1\%. Adding dense candidates raises correctness further to 47.3\%, compared with 34.0\% for GPT-5 reranking. This configuration also achieves the highest exact match on Natural Questions (54.8\%) and ConditionalQA (47.6\%), where it approaches the performance ceiling set by the gold passages (49.4\%).

\subsection{Learning a Retrieval Judge}
\label{sec:results:trained}

The retrieval framework also supports task-trained, open-weight decision models. We adapt Qwen3-8B using MultiDoc2Dial training data (\cref{sec:setup:train}). The resulting model achieves 60.7 R@1 on MultiDoc2Dial, exceeding the strongest published task-trained retriever by 5.7 points (\cref{tab:trained-judge}).
This demonstrates that the decision-based retrieval framework is not limited to the pretrained \jev{} model and can benefit from task-specific adaptation.

\begin{table}[h]
\centering
\small
\setlength{\tabcolsep}{1pt}
\caption{Retrieval with task-trained methods (\%).}
\label{tab:trained-judge}

\begin{tabular*}{\columnwidth}{
@{\extracolsep{\fill}}lrr@{}}
\toprule
& \multicolumn{2}{c}{MultiDoc2Dial} \\
\cmidrule(lr){2-3}
Method & R@1 & R@10 \\
\midrule
\multicolumn{3}{@{}l}{
\textit{Task-trained conventional retrieval}} \\
DPR \citep{feng2021multidoc2dial}\textsuperscript{p}
& 49.0 & 80.0 \\
SFRe2g \citep{sfre2g2026}\textsuperscript{p}
& 55.0 & 83.2 \\
\midrule
\multicolumn{3}{@{}l}{
\textit{Task-trained decision-based retrieval (ours)}} \\
\textbf{\method{} (8B trained)}
& 60.7 & 85.8 \\
\bottomrule
\end{tabular*}

\vspace{2pt}
\begin{minipage}{\columnwidth}
\footnotesize
\textsuperscript{p}Published results, not reruns.
\end{minipage}
\end{table}

\subsection{Latency and Cost}
\label{sec:results:cost}

\begin{table}[h]
\centering\small
\setlength{\tabcolsep}{1.5pt}
\caption{Median latency (seconds) and API cost per query. DecisionRAG rows are live, one query at a time, without reused answers; GPT-5 covers reranking only. Dense top-100 + Qwen3-RR is served locally on one NVIDIA RTX A6000. Qwen3-RR: Qwen3-Reranker-8B; NQ: Natural Questions.}
\label{tab:latency}
\begin{tabular}{@{}lrrrr@{}}
\toprule
& MD2D & CQA & MTRAG & NQ \\
\midrule
\multicolumn{5}{@{}l}{\textit{Median latency (s)}} \\
Dense top-100 + Qwen3-RR & 7.5 & 3.3 & 14.3 & 5.8 \\
Dense top-20 + GPT-5 & 11.8 & 14.4 & 12.2 & 15.0 \\
\textbf{\method{} (\jev{})} & 0.68 & 0.82 & 1.76 & 1.73 \\
\quad \textbf{+ dense candidates} & 0.70 & 0.85 & 1.73 & 1.64 \\
\midrule
\multicolumn{5}{@{}l}{\textit{API cost per query (¢)}} \\
Dense top-20 + GPT-5 & 1.40 & 1.20 & 1.90 & 1.40 \\
\textbf{\method{} (\jev{})} & 0.22 & 0.29 & 1.73 & 2.33 \\
\quad \textbf{+ dense candidates} & 0.23 & 0.29 & 1.75 & 2.33 \\
\bottomrule
\end{tabular}
\end{table}

Finally, we measure retrieval latency and API cost (\cref{tab:latency} and \cref{app:timing}). Under the measured serving configurations, \method{} completes retrieval in 0.68–1.76 seconds per query, compared with 3.3–14.3 seconds for dense retrieval with Qwen3-Reranker-8B on one GPU and 11.8–15.0 seconds for GPT-5 reranking.

API costs range from 0.22¢ per query on MultiDoc2Dial to 2.33¢ on Natural Questions. Despite the much larger Natural Questions corpus, its measured latency is comparable to MTRAG (1.73 versus 1.76 seconds), illustrating how hierarchical structures can reduce retrieval latency (\cref{eq:scaling}).
\section{Understanding DecisionRAG}
\label{sec:ablations}
\label{sec:ablation}

What makes decision-based retrieval effective? We examine how shared context sharpens relevance judgments, when multiple-choice ranking helps, and how navigation strategies and budgets trade off retrieval quality against efficiency.

\subsection{Scoring Candidates in a Shared Context}
\label{sec:ablations:shared}
\label{sec:ablation:context}
\label{sec:ablations:pool}

\begin{figure}[h]
    \centering
    \includegraphics[width=\columnwidth]{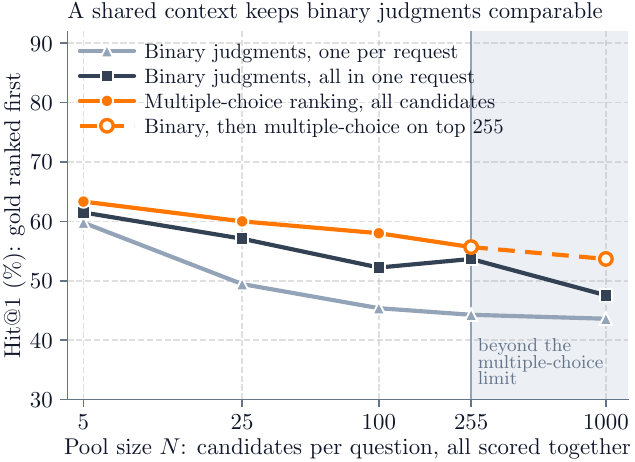}
    \caption{Hit@1 on Natural Questions as the candidate pool grows, with all $N$ candidates scored in one request (150 questions).}
    \label{fig:pool-size}
\end{figure}

A central design choice of \method{} is to judge candidates against a shared context rather than independently. For Natural Questions design questions, we build pools of 5 to 1,000 candidates from the gold paragraph and hard negatives retrieved by BM25, SPLADE-v3, and e5 (\cref{fig:pool-size}). Shared-context binary judgments consistently outperform scoring each candidate in a separate request, and multiple-choice ranking performs best within its 255-option limit. Beyond that limit, multiple-choice ranking of the binary top 255 retains much of this advantage: 53.7\% Hit@1 with 1,000 candidates versus 47.6\% for binary judgments alone.

The advantage carries over to the full pipeline: on MultiDoc2Dial, scoring the passages of all selected pages together rather than page by page raises R@1 from 36.7\% to 44.4\% (+7.7 points, 95\% CI: [3.3, 12.2]; \cref{app:context}).

\subsection{Request Size}
\label{sec:ablation:ctxsize}

In a shared context, each probability is formed in view of the alternatives: a candidate that would look relevant on its own receives a lower $P(\mathrm{Yes})$ when a clearly better one is present. The shared context thus puts the individual probabilities into perspective, making them comparable across candidates. We analyze how this calibration improves as the number of candidates in the context increases.

\begin{figure}[h]
    \centering
    \includegraphics[width=\columnwidth]{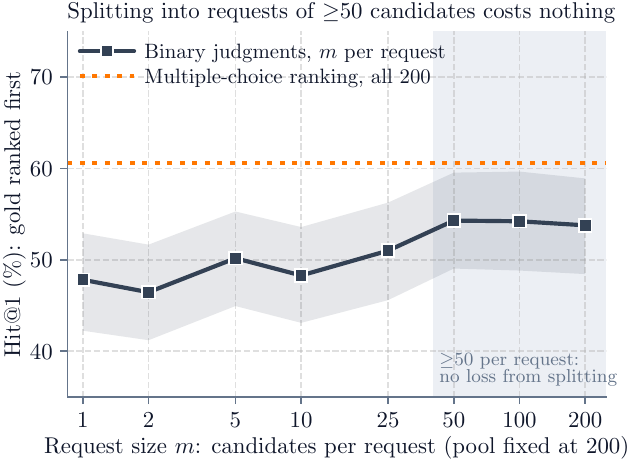}
    \caption{Hit@1 on Natural Questions when a fixed pool of 200 candidates is split into requests of $m$ candidates and ranked together (289 questions; band: 95\% interval).}
    \label{fig:request-size}
\end{figure}

To this end, we fix the pool at 200 candidates and split it into requests of varying sizes (\cref{fig:request-size}). Hit@1 increases from 47.8\% when each candidate is scored separately to 54.3\% with 50 candidates per request, with little further improvement for larger requests, so large pools can be split into parallel requests of about 50 candidates without loss. Multiple-choice ranking of all 200 candidates reaches 60.6\%, another 6.8 points above binary judgments of all 200 in one request (95\% CI: [2.5, 11.0]).

\subsection{Decisions versus Generation}
\label{sec:ablations:generative}
\label{app:generative}

Prior work also navigates hierarchies with language models, but lets a
generative LLM decide at each level via generating tokens \citep{chen2023memwalker, vectify2026pageindex, gupta2026lattice}. To isolate the effect of deciding by parallel
judgments instead of generation, we keep \method{}'s hierarchy and budgets
fixed and replace its decision model by GPT-5 (low reasoning effort), which
reads the candidates of each level and writes out the numbers of those to
keep, in the style of LATTICE~\citep{gupta2026lattice}. On ConditionalQA it selects 3 of the 652
pages and ranks their paragraphs; on Natural Questions it selects 150 of
the 1,000 clusters (when it lists fewer, we ask for the rest in the same
conversation), then 30 pages from their titles, and ranks the
paragraphs of the top pages as \method{} does.

\begin{table}[h]
\centering\small
\setlength{\tabcolsep}{4pt}
\caption{The same hierarchy, decided by generation versus
decisions. Latency: median seconds per query; cost: cents per query.
Natural Questions: the first 100 test questions of
\cref{tab:retrieval-summary}.}
\label{tab:generative}
\begin{tabular}{@{}lrrrr@{}}
\toprule
& Hit@1 & R@10 & s & ¢ \\
\midrule
\multicolumn{5}{@{}l}{\textit{ConditionalQA}} \\
GPT-5, generative & 73.8 & 73.6 & 28.2 & 6.1 \\
\textbf{\method{}} & 69.7 & 71.9 & 0.82 & 0.29 \\
\quad\textbf{+ dense candidates} & 73.4 & 76.3 & 0.85 & 0.29 \\
\midrule
\multicolumn{5}{@{}l}{\textit{Natural Questions}} \\
GPT-5, generative & 59.0 & 82.5 & 250.8 & 57.2 \\
\textbf{\method{}} & 65.0 & 86.5 & 1.73 & 2.33 \\
\quad\textbf{+ dense candidates} & 64.0 & 88.5 & 1.64 & 2.33 \\
\bottomrule
\end{tabular}
\end{table}

Generating the decisions token-by-token brings no consistent gain in accuracy (\cref{tab:generative}): GPT-5 leads \method{} by 4.1 Hit@1 points on ConditionalQA, a gap that adding dense candidates closes, and trails it by 6.0 points on Natural Questions, while being 34 to 145 times slower and 21 to 25 times more expensive, because it must read and reason over every candidate of a level before naming them sequentially token-by-token.

\subsection{Does Multiple-Choice Ranking Improve Retrieval?}
\label{sec:ablations:choice}
\label{sec:ablation:choice}

After the binary judgments, \method{} optionally applies multiple-choice ranking to the top candidates to refine the final ranking. We examine whether this additional step improves retrieval beyond parallel binary judgments alone.

On MultiDoc2Dial and ConditionalQA design data, we compare configurations with and without multiple-choice ranking while keeping the candidate pool and preceding binary judgments fixed. Multiple-choice ranking improves MultiDoc2Dial R@1 from 50.1\% to 51.3\% and ConditionalQA Hit@1 from 71.9\% to 76.5\% (\cref{tab:choice-ablation}). Both configurations include additional dense retrieval candidates.

\begin{table}[h]
\centering\small
\setlength{\tabcolsep}{3pt}
\caption{Multiple-choice ranking on design data (\%; MultiDoc2Dial: 1,832 turns, ConditionalQA: 285 questions). Multiple-choice ranking reorders the top 20 passages, with only these passages in its context (\cref{app:retrieval:choice}).}
\label{tab:choice-ablation}
\begin{tabular}{@{}llrr@{}}
\toprule
Dataset & Metric & \shortstack{Binary\\only} & \shortstack{+ Multiple-\\choice} \\
\midrule
MultiDoc2Dial & R@1 & 50.1 & 51.3 \\
ConditionalQA & Hit@1 & 71.9 & 76.5 \\
\bottomrule
\end{tabular}
\end{table}

The additional request takes 0.12--0.15 seconds and increases API cost by 4--33\%, making multiple-choice ranking a cheap final step that sharpens the top rank.

\subsection{Navigation versus Reranking}
\label{sec:ablations:reranking}

Do the gains come from navigating the hierarchy or from the decision model itself? We give \jev{} the top 100 passages of dense retrieval (Qwen3-Embedding-8B) and score them as \method{}'s last stage does: binary judgments in one shared context, followed by multiple-choice ranking of the top 20 (top five on Natural Questions). We compare this reranker with \method{}.

\begin{table}[h]
\centering\small
\setlength{\tabcolsep}{4pt}
\caption{\jev{} reranking the dense top 100 versus \method{} with dense candidates, both using the same dense index (\%). MD2D: MultiDoc2Dial, seeded sample of 1,000 validation turns; CQA: ConditionalQA; MTRAG: official labels; NQ: Natural Questions.}
\label{tab:navigation}
\begin{tabular}{@{}llrr@{}}
\toprule
& & \shortstack{Dense top-100\\+ \jev{} Rerank} & \shortstack{\method{}\\+ dense candidates} \\
\midrule
MD2D & R@1 & 49.9 & 51.2 \\
 & R@10 & 86.6 & 85.6 \\
CQA & Hit@1 & 65.7 & 73.4 \\
 & R@10 & 68.2 & 76.3 \\
MTRAG & R@5 & 38.3 & 38.7 \\
 & nDCG@5 & 36.8 & 35.7 \\
NQ & Hit@1 & 59.4 & 66.4 \\
 & nDCG@10 & 72.2 & 77.6 \\
\bottomrule
\end{tabular}
\end{table}

Navigating the hierarchy outperforms reranking the retriever's candidates (\cref{tab:navigation}): Hit@1 rises by 7.7 points on ConditionalQA (95\% CI: [2.6, 13.3]) and by 7.0 points on Natural Questions ([3.4, 10.8]), while MultiDoc2Dial and MTRAG are level. Even without any retrieval index, \method{} (\cref{tab:retrieval-summary}) improves Hit@1 over the reranker by 6.0 points on Natural Questions ([2.4, 9.8]) and by 4.0 points on ConditionalQA. This holds although the dense top 100 contains the gold evidence more often than the navigated candidates (Natural Questions: 98.4\% versus 94.0\%; ConditionalQA: 98.5\% versus 91.1\%): when both contain it, navigation ranks it first more often (70.1\% versus 61.9\% and 76.7\% versus 67.3\%), plausibly because it presents whole pages in reading order rather than isolated passages.

\subsection{Navigation Budget and Hierarchy}
\label{sec:ablations:budget}
\label{sec:ablation:coverage}

Finally, we examine how the navigation budget affects retrieval.

\begin{figure}[h]
    \centering
    \includegraphics[width=\columnwidth]{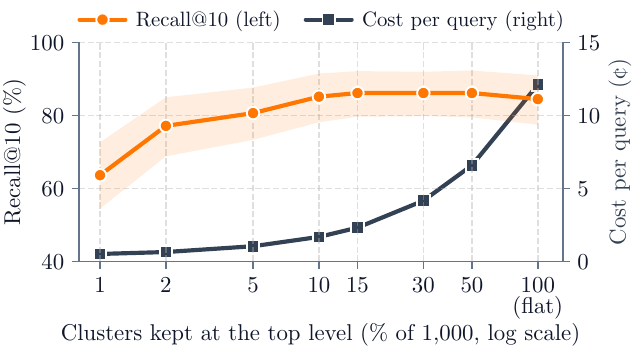}
    \caption{Recall@10 (left axis) and \jev{} cost per query (right axis) on Natural Questions as the top level keeps more and more clusters. First 100 test questions of \cref{tab:retrieval-summary}; band: 95\% interval.}
    \label{fig:cluster-share}
\end{figure}

Pruning at the top level is what keeps navigation cheap on large corpora, but it is also where evidence can be lost: a page in a discarded cluster can never be retrieved. We investigate how many clusters must be kept to match the recall of scoring every page directly, varying the share of the 1,000 Natural Questions clusters that \method{} keeps up to flat scoring of all 108,626 page titles (\cref{fig:cluster-share}). Recall@10 rises steeply up to about 10\% of the clusters and then stays level: keeping 15\% matches flat scoring (86.2\% versus 84.5\%) at a fifth of the cost (2.3¢ versus 12.1¢ per query) and a third of the latency (1.7 versus 4.5 seconds). Below 5\%, the cluster holding the gold page is often discarded, and recall@10 falls to 63.7\% at 1\%.

The same trade-off appears one level down, in the page budget. On MultiDoc2Dial, we increase the number of newly retained pages from three to ten, keeping the remaining configuration fixed. R@10 improves from 81.0\% to 83.6\%, while R@1 increases only from 46.9\% to 47.4\% (\cref{tab:budget}). The larger budget increases API cost by 83\% and median latency by 28\%. Exploring more pages therefore primarily improves evidence coverage, with diminishing returns for the top-ranked passage.

\begin{table}[h]
\centering\small
\setlength{\tabcolsep}{4pt}
\caption{Page budget on MultiDoc2Dial (validation, with multiple-choice ranking).}
\label{tab:budget}
\begin{tabular}{@{}crrrr@{}}
\toprule
\shortstack{New\\pages} & \shortstack{R@1\\(\%)} & \shortstack{R@10\\(\%)} & \shortstack{API cost\\(¢/query)} & \shortstack{Median\\time (s)} \\
\midrule
3 & 46.9 & 81.0 & 0.12 & 0.53 \\
10 & 47.4 & 83.6 & 0.22 & 0.68 \\
\bottomrule
\end{tabular}
\end{table}

\section{Limitations}
\label{sec:limitations}

\method{} relies on a corpus hierarchy whose structure and node descriptions
allow the decision model to identify relevant branches. Poorly organized
documents or uninformative descriptions can therefore hurt retrieval quality.
Moreover, navigation prunes branches irreversibly, so relevant passages may be
missed if their ancestors are discarded; we show that larger exploration budgets and
complementary dense retrieval can mitigate this.

\section{Conclusion}
\label{sec:conclusion}

We introduced \method{}, which brings the capabilities of large language models into retrieval by navigating corpus hierarchies with non-generative decisions. At each level, candidates are evaluated jointly in a shared context, allowing their relevance to be judged in view of one another. By progressively narrowing the search space, \method{} retrieves relevant evidence with a number of sequential decisions that grows only logarithmically with corpus size under a balanced hierarchy. As decision models improve, retrieval can inherit their capabilities directly.

\ifarxiv
\section*{Acknowledgements}

This work was supported by the National Science Foundation (NSF) under grants OAC-2118310 and IIS-2530143, through the AI Research Institutes program (Award No.\ DMR-2433348), by NewYork-Presbyterian for the NYP-Cornell Cardiovascular AI Collaboration, and by arXiv.
\fi

\ifarxiv\else\nocite{*}\fi

\bibliography{sections/References,sections/additional_references}

@inproceedings{chen2023densex,
  author       = {Tong Chen and
                  Hongwei Wang and
                  Sihao Chen and
                  Wenhao Yu and
                  Kaixin Ma and
                  Xinran Zhao and
                  Hongming Zhang and
                  Dong Yu},
  title        = {Dense {X} Retrieval: What Retrieval Granularity Should We Use?},
  booktitle    = {{EMNLP}},
  pages        = {15159--15177},
  publisher    = {Association for Computational Linguistics},
  year         = {2024}
}

@inproceedings{lewis2020rag,
  author       = {Patrick Lewis and
                  Ethan Perez and
                  Aleksandra Piktus and
                  Fabio Petroni and
                  Vladimir Karpukhin and
                  Naman Goyal and
                  Heinrich K{\"{u}}ttler and
                  Mike Lewis and
                  Wen{-}tau Yih and
                  Tim Rockt{\"{a}}schel and
                  Sebastian Riedel and
                  Douwe Kiela},
  title        = {Retrieval-Augmented Generation for Knowledge-Intensive {NLP} Tasks},
  booktitle    = {NeurIPS},
  year         = {2020}
}

@inproceedings{glass2022re2g,
  author       = {Michael R. Glass and
                  Gaetano Rossiello and
                  Md. Faisal Mahbub Chowdhury and
                  Ankita Naik and
                  Pengshan Cai and
                  Alfio Gliozzo},
  title        = {Re2G: Retrieve, Rerank, Generate},
  booktitle    = {{NAACL-HLT}},
  pages        = {2701--2715},
  publisher    = {Association for Computational Linguistics},
  year         = {2022}
}

@inproceedings{asai2024selfrag,
  author       = {Akari Asai and
                  Zeqiu Wu and
                  Yizhong Wang and
                  Avirup Sil and
                  Hannaneh Hajishirzi},
  title        = {Self-RAG: Learning to Retrieve, Generate, and Critique through
                  Self-Reflection},
  booktitle    = {{ICLR}},
  publisher    = {OpenReview.net},
  year         = {2024}
}

@article{yan2024crag,
  author       = {Shi{-}Qi Yan and
                  Jia{-}Chen Gu and
                  Yun Zhu and
                  Zhen{-}Hua Ling},
  title        = {Corrective Retrieval Augmented Generation},
  journal      = {CoRR},
  volume       = {abs/2401.15884},
  year         = {2024}
}

@inproceedings{xu2024recomp,
  author       = {Fangyuan Xu and
                  Weijia Shi and
                  Eunsol Choi},
  title        = {{RECOMP:} Improving Retrieval-Augmented LMs with Context Compression and
                  Selective Augmentation},
  booktitle    = {{ICLR}},
  publisher    = {OpenReview.net},
  year         = {2024}
}

@inproceedings{lee2025setr,
  author       = {Dahyun Lee and
                  Yongrae Jo and
                  Haeju Park and
                  Moontae Lee},
  title        = {Shifting from Ranking to Set Selection for Retrieval Augmented
                  Generation},
  booktitle    = {{ACL} {(1)}},
  pages        = {17606--17619},
  publisher    = {Association for Computational Linguistics},
  year         = {2025}
}

@inproceedings{sarthi2024raptor,
  author       = {Parth Sarthi and
                  Salman Abdullah and
                  Aditi Tuli and
                  Shubh Khanna and
                  Anna Goldie and
                  Christopher D. Manning},
  title        = {{RAPTOR:} Recursive Abstractive Processing for Tree-Organized Retrieval},
  booktitle    = {{ICLR}},
  publisher    = {OpenReview.net},
  year         = {2024}
}

@article{chen2023memwalker,
  author       = {Howard Chen and
                  Ramakanth Pasunuru and
                  Jason Weston and
                  Asli Celikyilmaz},
  title        = {Walking Down the Memory Maze: Beyond Context Limit through Interactive
                  Reading},
  journal      = {CoRR},
  volume       = {abs/2310.05029},
  year         = {2023}
}

@article{sun2021dochopper,
  author       = {Haitian Sun and
                  William W. Cohen and
                  Ruslan Salakhutdinov},
  title        = {End-to-End Multihop Retrieval for Compositional Question Answering over
                  Long Documents},
  journal      = {CoRR},
  volume       = {abs/2106.00200},
  year         = {2021}
}

@inproceedings{wang2024dapr,
  author       = {Kexin Wang and
                  Nils Reimers and
                  Iryna Gurevych},
  title        = {{DAPR:} {A} Benchmark on Document-Aware Passage Retrieval},
  booktitle    = {{ACL} {(1)}},
  pages        = {4313--4330},
  publisher    = {Association for Computational Linguistics},
  year         = {2024}
}

@article{sfre2g2026,
  author       = {Yeqin Zhang and
                  Haomin Fu and
                  Xujie Zhang and
                  Cam{-}Tu Nguyen},
  title        = {Structures Facilitate Retrieve, Rerank, and Generate},
  journal      = {CoRR},
  volume       = {abs/2606.03247},
  year         = {2026}
}

@inproceedings{geva2018docqn,
  author       = {Mor Geva and
                  Jonathan Berant},
  title        = {Learning to Search in Long Documents Using Document Structure},
  booktitle    = {{COLING}},
  pages        = {161--176},
  publisher    = {Association for Computational Linguistics},
  year         = {2018}
}

@inproceedings{xu2025rdr2,
  author       = {Lingnan Xu and
                  Chong Feng and
                  Kaiyuan Zhang and
                  Liu Zhengyong and
                  Wenqiang Xu and
                  Fanqing Meng},
  title        = {Equipping Retrieval-Augmented Large Language Models with Document
                  Structure Awareness},
  booktitle    = {{EMNLP} (Findings)},
  pages        = {24608--24631},
  publisher    = {Association for Computational Linguistics},
  year         = {2025}
}

@article{jayraj2026navtree,
  title = {Tree Navigation Without {LLM} Summaries: A Matched-Cost Study of Hierarchical Retrieval for Long-Document {QA}},
  author = {Jayraj, Priyank and Goyal, Poonam and Goyal, Navneet},
  year = {2026},
  journal = {arXiv preprint arXiv:2610.06902},
  url = {https://arxiv.org/abs/2610.06902}
}

@misc{vectify2026pageindex,
  title = {{PageIndex}: Document Index for Reasoning-Based {RAG}},
  author = {{VectifyAI}},
  year = {2025},
  howpublished = {Software repository},
  url = {https://github.com/VectifyAI/PageIndex}
}

@inproceedings{sun2023rankgpt,
  author       = {Weiwei Sun and
                  Lingyong Yan and
                  Xinyu Ma and
                  Shuaiqiang Wang and
                  Pengjie Ren and
                  Zhumin Chen and
                  Dawei Yin and
                  Zhaochun Ren},
  title        = {Is ChatGPT Good at Search? Investigating Large Language Models as
                  Re-Ranking Agents},
  booktitle    = {{EMNLP}},
  pages        = {14918--14937},
  publisher    = {Association for Computational Linguistics},
  year         = {2023}
}

@article{ma2023listwise,
  author       = {Xueguang Ma and
                  Xinyu Zhang and
                  Ronak Pradeep and
                  Jimmy Lin},
  title        = {Zero-Shot Listwise Document Reranking with a Large Language Model},
  journal      = {CoRR},
  volume       = {abs/2305.02156},
  year         = {2023}
}

@article{pradeep2023rankzephyr,
  author       = {Ronak Pradeep and
                  Sahel Sharifymoghaddam and
                  Jimmy Lin},
  title        = {RankZephyr: Effective and Robust Zero-Shot Listwise Reranking is a
                  Breeze!},
  journal      = {CoRR},
  volume       = {abs/2312.02724},
  year         = {2023}
}

@article{li2026llmasjev,
  title = {{LLM-as-Jev}: {LLMs} Are Already {Jev-Style} Decision Models -- When and How to {Fine-Tune} Them},
  author = {Li, Yinheng and Wagle, Justin},
  year = {2026},
  journal = {arXiv preprint arXiv:2610.02076},
  url = {https://arxiv.org/abs/2610.02076}
}

@inproceedings{nogueira2020monot5,
  author       = {Rodrigo Nogueira and
                  Zhiying Jiang and
                  Ronak Pradeep and
                  Jimmy Lin},
  title        = {Document Ranking with a Pretrained Sequence-to-Sequence Model},
  booktitle    = {{EMNLP} (Findings)},
  series       = {Findings of {ACL}},
  volume       = {EMNLP 2020},
  pages        = {708--718},
  publisher    = {Association for Computational Linguistics},
  year         = {2020}
}

@inproceedings{zhuang2023rankt5,
  author       = {Honglei Zhuang and
                  Zhen Qin and
                  Rolf Jagerman and
                  Kai Hui and
                  Ji Ma and
                  Jing Lu and
                  Jianmo Ni and
                  Xuanhui Wang and
                  Michael Bendersky},
  title        = {RankT5: Fine-Tuning {T5} for Text Ranking with Ranking Losses},
  booktitle    = {{SIGIR}},
  pages        = {2308--2313},
  publisher    = {{ACM}},
  year         = {2023}
}

@inproceedings{reddy2024first,
  author       = {Revanth Gangi Reddy and
                  JaeHyeok Doo and
                  Yifei Xu and
                  Md. Arafat Sultan and
                  Deevya Swain and
                  Avirup Sil and
                  Heng Ji},
  title        = {{FIRST:} Faster Improved Listwise Reranking with Single Token Decoding},
  booktitle    = {{EMNLP}},
  pages        = {8642--8652},
  publisher    = {Association for Computational Linguistics},
  year         = {2024}
}

@inproceedings{chen2025icr,
  author       = {Shijie Chen and
                  Bernal Jimenez Gutierrez and
                  Yu Su},
  title        = {Attention in Large Language Models Yields Efficient Zero-Shot Re-Rankers},
  booktitle    = {{ICLR}},
  publisher    = {OpenReview.net},
  year         = {2025}
}

@inproceedings{zhuang2024setwise,
  author       = {Shengyao Zhuang and
                  Honglei Zhuang and
                  Bevan Koopman and
                  Guido Zuccon},
  title        = {A Setwise Approach for Effective and Highly Efficient Zero-shot Ranking
                  with Large Language Models},
  booktitle    = {{SIGIR}},
  pages        = {38--47},
  publisher    = {{ACM}},
  year         = {2024}
}

@inproceedings{yoon2024listt5,
  author       = {Soyoung Yoon and
                  Eunbi Choi and
                  Jiyeon Kim and
                  Hyeongu Yun and
                  Yireun Kim and
                  Seung{-}won Hwang},
  title        = {ListT5: Listwise Reranking with Fusion-in-Decoder Improves Zero-shot
                  Retrieval},
  booktitle    = {{ACL} {(1)}},
  pages        = {2287--2308},
  publisher    = {Association for Computational Linguistics},
  year         = {2024}
}

@article{li2026wholepool,
  author       = {Hang Li and
                  Chuting Yu and
                  Teerapong Leelanupab and
                  Bevan Koopman and
                  Guido Zuccon},
  title        = {Whole-Pool Setwise Reranking with Long-Context Language Models},
  journal      = {CoRR},
  volume       = {abs/2606.01782},
  year         = {2026}
}

@inproceedings{reimers2021curse,
  author       = {Nils Reimers and
                  Iryna Gurevych},
  title        = {The Curse of Dense Low-Dimensional Information Retrieval for Large Index
                  Sizes},
  booktitle    = {{ACL/IJCNLP} {(2)}},
  pages        = {605--611},
  publisher    = {Association for Computational Linguistics},
  year         = {2021}
}

@inproceedings{pradeep2023generative,
  author       = {Ronak Pradeep and
                  Kai Hui and
                  Jai Gupta and
                  {\'{A}}d{\'{a}}m D. Lelkes and
                  Honglei Zhuang and
                  Jimmy Lin and
                  Donald Metzler and
                  Vinh Q. Tran},
  title        = {How Does Generative Retrieval Scale to Millions of Passages?},
  booktitle    = {{EMNLP}},
  pages        = {1305--1321},
  publisher    = {Association for Computational Linguistics},
  year         = {2023}
}

@article{johnson2019faiss,
  author       = {Jeff Johnson and
                  Matthijs Douze and
                  Herv{\'{e}} J{\'{e}}gou},
  title        = {Billion-Scale Similarity Search with GPUs},
  journal      = {{IEEE} Trans. Big Data},
  volume       = {7},
  number       = {3},
  pages        = {535--547},
  year         = {2021}
}

@inproceedings{chen2021spann,
  author       = {Qi Chen and
                  Bing Zhao and
                  Haidong Wang and
                  Mingqin Li and
                  Chuanjie Liu and
                  Zengzhong Li and
                  Mao Yang and
                  Jingdong Wang},
  title        = {{SPANN:} Highly-efficient Billion-scale Approximate Nearest Neighborhood
                  Search},
  booktitle    = {NeurIPS},
  pages        = {5199--5212},
  year         = {2021}
}

@article{kulkarni2015selective,
  author       = {Anagha Kulkarni and
                  Jamie Callan},
  title        = {Selective Search: Efficient and Effective Search of Large Textual
                  Collections},
  journal      = {{ACM} Trans. Inf. Syst.},
  volume       = {33},
  number       = {4},
  pages        = {17:1--17:33},
  year         = {2015}
}

@inproceedings{dai2016querybiased,
  author       = {Zhuyun Dai and
                  Chenyan Xiong and
                  Jamie Callan},
  title        = {Query-Biased Partitioning for Selective Search},
  booktitle    = {{CIKM}},
  pages        = {1119--1128},
  publisher    = {{ACM}},
  year         = {2016}
}

@inproceedings{dong2020neurallsh,
  author       = {Yihe Dong and
                  Piotr Indyk and
                  Ilya P. Razenshteyn and
                  Tal Wagner},
  title        = {Learning Space Partitions for Nearest Neighbor Search},
  booktitle    = {{ICLR}},
  publisher    = {OpenReview.net},
  year         = {2020}
}

@inproceedings{gupta2022bliss,
  author       = {Gaurav Gupta and
                  Tharun Medini and
                  Anshumali Shrivastava and
                  Alexander J. Smola},
  title        = {{BLISS:} {A} Billion scale Index using Iterative Re-partitioning},
  booktitle    = {{KDD}},
  pages        = {486--495},
  publisher    = {{ACM}},
  year         = {2022}
}

@inproceedings{lv2007multiprobe,
  author       = {Qin Lv and
                  William Josephson and
                  Zhe Wang and
                  Moses Charikar and
                  Kai Li},
  title        = {Multi-Probe {LSH:} Efficient Indexing for High-Dimensional Similarity
                  Search},
  booktitle    = {{VLDB}},
  pages        = {950--961},
  publisher    = {{ACM}},
  year         = {2007}
}

@inproceedings{aly2013taily,
  author       = {Robin Aly and
                  Djoerd Hiemstra and
                  Thomas Demeester},
  title        = {Taily: shard selection using the tail of score distributions},
  booktitle    = {{SIGIR}},
  pages        = {673--682},
  publisher    = {{ACM}},
  year         = {2013}
}

@inproceedings{kulkarni2015shrkc,
  author       = {Anagha Kulkarni},
  title        = {ShRkC: Shard Rank Cutoff Prediction for Selective Search},
  booktitle    = {{SPIRE}},
  series       = {Lecture Notes in Computer Science},
  volume       = {9309},
  pages        = {337--349},
  publisher    = {Springer},
  year         = {2015}
}

@inproceedings{li2020termination,
  author       = {Conglong Li and
                  Minjia Zhang and
                  David G. Andersen and
                  Yuxiong He},
  title        = {Improving Approximate Nearest Neighbor Search through Learned Adaptive
                  Early Termination},
  booktitle    = {{SIGMOD} Conference},
  pages        = {2539--2554},
  publisher    = {{ACM}},
  year         = {2020}
}

@inproceedings{athanasiou2026ails,
  author       = {Dimosthenis Athanasiou and
                  Maria Lymperaiou and
                  Giorgos Filandrianos and
                  Athanasios Voulodimos and
                  Giorgos Stamou},
  title        = {{AILS-NTUA} at SemEval-2026 Task 8: Evaluating Multi-Turn {RAG}
                  Conversations},
  booktitle    = {SemEval@ACL},
  pages        = {1340--1365},
  publisher    = {Association for Computational Linguistics},
  year         = {2026}
}

@inproceedings{lupart2026uva,
  author       = {Simon Lupart and
                  Kidist Amde Mekonnen and
                  Zahra Abbasiantaeb and
                  Mohammad Aliannejadi},
  title        = {uva-irlab-conv at SemEval-2026 Task 8: Multi-Turn {RAG} with Learned
                  Sparse Retrieval and Listwise Reranking},
  booktitle    = {SemEval@ACL},
  pages        = {2972--2980},
  publisher    = {Association for Computational Linguistics},
  year         = {2026}
}

@inproceedings{yu2020fewshotcqr,
  author       = {Shi Yu and
                  Jiahua Liu and
                  Jingqin Yang and
                  Chenyan Xiong and
                  Paul N. Bennett and
                  Jianfeng Gao and
                  Zhiyuan Liu},
  title        = {Few-Shot Generative Conversational Query Rewriting},
  booktitle    = {{SIGIR}},
  pages        = {1933--1936},
  publisher    = {{ACM}},
  year         = {2020}
}

@inproceedings{mo2023convgqr,
  author       = {Fengran Mo and
                  Kelong Mao and
                  Yutao Zhu and
                  Yihong Wu and
                  Kaiyu Huang and
                  Jian{-}Yun Nie},
  title        = {ConvGQR: Generative Query Reformulation for Conversational Search},
  booktitle    = {{ACL} {(1)}},
  pages        = {4998--5012},
  publisher    = {Association for Computational Linguistics},
  year         = {2023}
}

@inproceedings{mao2023llm4cs,
  author       = {Kelong Mao and
                  Zhicheng Dou and
                  Fengran Mo and
                  Jiewen Hou and
                  Haonan Chen and
                  Hongjin Qian},
  title        = {Large Language Models Know Your Contextual Search Intent: {A} Prompting
                  Framework for Conversational Search},
  booktitle    = {{EMNLP} (Findings)},
  series       = {Findings of {ACL}},
  volume       = {EMNLP 2023},
  pages        = {1211--1225},
  publisher    = {Association for Computational Linguistics},
  year         = {2023}
}

@inproceedings{formal2021splade,
  author       = {Thibault Formal and
                  Benjamin Piwowarski and
                  St{\'{e}}phane Clinchant},
  title        = {{SPLADE:} Sparse Lexical and Expansion Model for First Stage Ranking},
  booktitle    = {{SIGIR}},
  pages        = {2288--2292},
  publisher    = {{ACM}},
  year         = {2021}
}

@article{formal2021spladev2,
  author       = {Thibault Formal and
                  Carlos Lassance and
                  Benjamin Piwowarski and
                  St{\'{e}}phane Clinchant},
  title        = {{SPLADE} v2: Sparse Lexical and Expansion Model for Information
                  Retrieval},
  journal      = {CoRR},
  volume       = {abs/2109.10086},
  year         = {2021}
}

@inproceedings{formal2022spladepp,
  author       = {Thibault Formal and
                  Carlos Lassance and
                  Benjamin Piwowarski and
                  St{\'{e}}phane Clinchant},
  title        = {From Distillation to Hard Negative Sampling: Making Sparse Neural {IR}
                  Models More Effective},
  booktitle    = {{SIGIR}},
  pages        = {2353--2359},
  publisher    = {{ACM}},
  year         = {2022}
}

@article{lassance2024spladev3,
  author       = {Carlos Lassance and
                  Herv{\'{e}} D{\'{e}}jean and
                  Thibault Formal and
                  St{\'{e}}phane Clinchant},
  title        = {SPLADE-v3: New baselines for {SPLADE}},
  journal      = {CoRR},
  volume       = {abs/2403.06789},
  year         = {2024}
}

@inproceedings{karpukhin2020dpr,
  author       = {Vladimir Karpukhin and
                  Barlas Oguz and
                  Sewon Min and
                  Patrick Lewis and
                  Ledell Wu and
                  Sergey Edunov and
                  Danqi Chen and
                  Wen{-}tau Yih},
  title        = {Dense Passage Retrieval for Open-Domain Question Answering},
  booktitle    = {{EMNLP} {(1)}},
  pages        = {6769--6781},
  publisher    = {Association for Computational Linguistics},
  year         = {2020}
}

@inproceedings{lin2023dragon,
  author       = {Sheng{-}Chieh Lin and
                  Akari Asai and
                  Minghan Li and
                  Barlas Oguz and
                  Jimmy Lin and
                  Yashar Mehdad and
                  Wen{-}tau Yih and
                  Xilun Chen},
  title        = {How to Train Your Dragon: Diverse Augmentation Towards Generalizable
                  Dense Retrieval},
  booktitle    = {{EMNLP} (Findings)},
  series       = {Findings of {ACL}},
  volume       = {EMNLP 2023},
  pages        = {6385--6400},
  publisher    = {Association for Computational Linguistics},
  year         = {2023}
}

@inproceedings{santhanam2022colbertv2,
  author       = {Keshav Santhanam and
                  Omar Khattab and
                  Jon Saad{-}Falcon and
                  Christopher Potts and
                  Matei Zaharia},
  title        = {ColBERTv2: Effective and Efficient Retrieval via Lightweight Late
                  Interaction},
  booktitle    = {{NAACL-HLT}},
  pages        = {3715--3734},
  publisher    = {Association for Computational Linguistics},
  year         = {2022}
}

@article{zhang2025qwen3embedding,
  author       = {Yanzhao Zhang and
                  Mingxin Li and
                  Dingkun Long and
                  Xin Zhang and
                  Huan Lin and
                  Baosong Yang and
                  Pengjun Xie and
                  An Yang and
                  Dayiheng Liu and
                  Junyang Lin and
                  Fei Huang and
                  Jingren Zhou},
  title        = {Qwen3 Embedding: Advancing Text Embedding and Reranking Through
                  Foundation Models},
  journal      = {CoRR},
  volume       = {abs/2506.05176},
  year         = {2025}
}

@inproceedings{chen2024bgem3,
  author       = {Jianlyu Chen and
                  Shitao Xiao and
                  Peitian Zhang and
                  Kun Luo and
                  Defu Lian and
                  Zheng Liu},
  title        = {M3-Embedding: Multi-Linguality, Multi-Functionality, Multi-Granularity
                  Text Embeddings Through Self-Knowledge Distillation},
  booktitle    = {{ACL} (Findings)},
  series       = {Findings of {ACL}},
  volume       = {ACL 2024},
  pages        = {2318--2335},
  publisher    = {Association for Computational Linguistics},
  year         = {2024}
}

@article{robertson2009bm25,
  author       = {Stephen E. Robertson and
                  Hugo Zaragoza},
  title        = {The Probabilistic Relevance Framework: {BM25} and Beyond},
  journal      = {Found. Trends Inf. Retr.},
  volume       = {3},
  number       = {4},
  pages        = {333--389},
  year         = {2009}
}

@misc{elastic2023elser,
  title = {{ELSER}: {E}lastic {L}earned {S}parse {E}ncode{R}},
  author = {{Elastic}},
  year = {2023},
  howpublished = {\url{https://www.elastic.co/guide/en/machine-learning/current/ml-nlp-elser.html}}
}

@misc{typesafe2026jev,
  title = {{J}ev 1.13: a decision model},
  author = {{TypeSafe AI}},
  year = {2026},
  howpublished = {\url{https://openrouter.ai/typesafe/jev-1.13}}
}

@misc{openai2026decisions,
  title = {{GPT-6} {L}una decisions model},
  author = {{OpenAI}},
  year = {2026},
  howpublished = {OpenRouter model \texttt{openai/gpt-6-luna-decisions}, snapshot 2026-10-06}
}

@misc{anthropic2026opus,
  title = {{C}laude {O}pus 5.5},
  author = {{Anthropic}},
  year = {2026},
  howpublished = {\url{https://www.anthropic.com/claude}}
}

@article{katsis2025mtrag,
  author       = {Yannis Katsis and
                  Sara Rosenthal and
                  Kshitij Fadnis and
                  Chulaka Gunasekara and
                  Young{-}Suk Lee and
                  Lucian Popa and
                  Vraj Shah and
                  Huaiyu Zhu and
                  Danish Contractor and
                  Marina Danilevsky},
  title        = {mtRAG: {A} Multi-Turn Conversational Benchmark for Evaluating
                  Retrieval-Augmented Generation Systems},
  journal      = {Trans. Assoc. Comput. Linguistics},
  volume       = {13},
  pages        = {784--808},
  year         = {2025}
}

@inproceedings{rosenthal2026mtragun,
  author       = {Sara Rosenthal and
                  Yannis Katsis and
                  Vraj Shah and
                  Lihong He and
                  Lucian Popa and
                  Marina Danilevsky},
  title        = {{MTRAG-UN:} {A} Benchmark for Open Challenges in Multi-Turn {RAG}
                  Conversations},
  booktitle    = {{ACL} (Findings)},
  pages        = {10363--10369},
  publisher    = {Association for Computational Linguistics},
  year         = {2026}
}

@article{trivedi2022musique,
  author       = {Harsh Trivedi and
                  Niranjan Balasubramanian and
                  Tushar Khot and
                  Ashish Sabharwal},
  title        = {MuSiQue: Multihop Questions via Single-hop Question Composition},
  journal      = {Trans. Assoc. Comput. Linguistics},
  volume       = {10},
  pages        = {539--554},
  year         = {2022}
}

@inproceedings{feng2021multidoc2dial,
  author       = {Song Feng and
                  Siva Sankalp Patel and
                  Hui Wan and
                  Sachindra Joshi},
  title        = {MultiDoc2Dial: Modeling Dialogues Grounded in Multiple Documents},
  booktitle    = {{EMNLP} {(1)}},
  pages        = {6162--6176},
  publisher    = {Association for Computational Linguistics},
  year         = {2021}
}

@inproceedings{sun2022conditionalqa,
  author       = {Haitian Sun and
                  William W. Cohen and
                  Ruslan Salakhutdinov},
  title        = {ConditionalQA: {A} Complex Reading Comprehension Dataset with
                  Conditional Answers},
  booktitle    = {{ACL} {(1)}},
  pages        = {3627--3637},
  publisher    = {Association for Computational Linguistics},
  year         = {2022}
}

@inproceedings{rosenthal2026semeval,
  author       = {Sara Rosenthal and
                  Vraj Shah and
                  Yannis Katsis and
                  Marina Danilevsky},
  title        = {SemEval-2026 Task 8: MTRAGEval: Evaluating Multi-Turn {RAG}
                  Conversations},
  booktitle    = {SemEval@ACL},
  pages        = {3673--3690},
  publisher    = {Association for Computational Linguistics},
  year         = {2026}
}

@inproceedings{thomas2024llmjudges,
  author       = {Paul Thomas and
                  Seth Spielman and
                  Nick Craswell and
                  Bhaskar Mitra},
  title        = {Large Language Models can Accurately Predict Searcher Preferences},
  booktitle    = {{SIGIR}},
  pages        = {1930--1940},
  publisher    = {{ACM}},
  year         = {2024}
}

@article{upadhyay2024umbrela,
  author       = {Shivani Upadhyay and
                  Ronak Pradeep and
                  Nandan Thakur and
                  Nick Craswell and
                  Jimmy Lin},
  title        = {{UMBRELA:} UMbrela is the (Open-Source Reproduction of the) Bing
                  RELevance Assessor},
  journal      = {CoRR},
  volume       = {abs/2406.06519},
  year         = {2024}
}

@inproceedings{faggioli2023perspectives,
  author       = {Guglielmo Faggioli and
                  Laura Dietz and
                  Charles L. A. Clarke and
                  Gianluca Demartini and
                  Matthias Hagen and
                  Claudia Hauff and
                  Noriko Kando and
                  Evangelos Kanoulas and
                  Martin Potthast and
                  Benno Stein and
                  Henning Wachsmuth},
  title        = {Perspectives on Large Language Models for Relevance Judgment},
  booktitle    = {{ICTIR}},
  pages        = {39--50},
  publisher    = {{ACM}},
  year         = {2023}
}

@article{buckley2007pooling,
  author       = {Chris Buckley and
                  Darrin Dimmick and
                  Ian Soboroff and
                  Ellen M. Voorhees},
  title        = {Bias and the limits of pooling for large collections},
  journal      = {Inf. Retr.},
  volume       = {10},
  number       = {6},
  pages        = {491--508},
  year         = {2007}
}

@inproceedings{zobel1998reliable,
  author       = {Justin Zobel},
  title        = {How Reliable Are the Results of Large-Scale Information Retrieval
                  Experiments?},
  booktitle    = {{SIGIR}},
  pages        = {307--314},
  publisher    = {{ACM}},
  year         = {1998}
}

@inproceedings{thakur2021beir,
  author       = {Nandan Thakur and
                  Nils Reimers and
                  Andreas R{\"{u}}ckl{\'{e}} and
                  Abhishek Srivastava and
                  Iryna Gurevych},
  title        = {{BEIR:} {A} Heterogeneous Benchmark for Zero-shot Evaluation of
                  Information Retrieval Models},
  booktitle    = {NeurIPS Datasets and Benchmarks},
  year         = {2021}
}

@article{edge2024graphrag,
  author       = {Darren Edge and
                  Ha Trinh and
                  Newman Cheng and
                  Joshua Bradley and
                  Alex Chao and
                  Apurva Mody and
                  Steven Truitt and
                  Jonathan Larson},
  title        = {From Local to Global: {A} Graph {RAG} Approach to Query-Focused
                  Summarization},
  journal      = {CoRR},
  volume       = {abs/2404.16130},
  year         = {2024}
}

@article{sun2022treasoner,
  author       = {Haitian Sun and
                  William W. Cohen and
                  Ruslan Salakhutdinov},
  title        = {Reasoning over Logically Interacted Conditions for Question Answering},
  journal      = {CoRR},
  volume       = {abs/2205.12898},
  year         = {2022}
}

@inproceedings{ainslie2020etc,
  author       = {Joshua Ainslie and
                  Santiago Onta{\~{n}}{\'{o}}n and
                  Chris Alberti and
                  Vaclav Cvicek and
                  Zachary Fisher and
                  Philip Pham and
                  Anirudh Ravula and
                  Sumit Sanghai and
                  Qifan Wang and
                  Li Yang},
  title        = {{ETC:} Encoding Long and Structured Inputs in Transformers},
  booktitle    = {{EMNLP} {(1)}},
  pages        = {268--284},
  publisher    = {Association for Computational Linguistics},
  year         = {2020}
}

@inproceedings{tay2022dsi,
  author       = {Yi Tay and
                  Vinh Tran and
                  Mostafa Dehghani and
                  Jianmo Ni and
                  Dara Bahri and
                  Harsh Mehta and
                  Zhen Qin and
                  Kai Hui and
                  Zhe Zhao and
                  Jai Prakash Gupta and
                  Tal Schuster and
                  William W. Cohen and
                  Donald Metzler},
  title        = {Transformer Memory as a Differentiable Search Index},
  booktitle    = {NeurIPS},
  year         = {2022}
}

@inproceedings{zhang2024beamretrieval,
  author       = {Jiahao Zhang and
                  Haiyang Zhang and
                  Dongmei Zhang and
                  Yong Liu and
                  Shen Huang},
  title        = {End-to-End Beam Retrieval for Multi-Hop Question Answering},
  booktitle    = {{NAACL-HLT}},
  pages        = {1718--1731},
  publisher    = {Association for Computational Linguistics},
  year         = {2024}
}

@inproceedings{feng2020doc2dial,
  author       = {Song Feng and
                  Hui Wan and
                  R. Chulaka Gunasekara and
                  Siva Sankalp Patel and
                  Sachindra Joshi and
                  Luis A. Lastras},
  title        = {doc2dial: {A} Goal-Oriented Document-Grounded Dialogue Dataset},
  booktitle    = {{EMNLP} {(1)}},
  pages        = {8118--8128},
  publisher    = {Association for Computational Linguistics},
  year         = {2020}
}

@inproceedings{dasigi2021qasper,
  author       = {Pradeep Dasigi and
                  Kyle Lo and
                  Iz Beltagy and
                  Arman Cohan and
                  Noah A. Smith and
                  Matt Gardner},
  title        = {A Dataset of Information-Seeking Questions and Answers Anchored in
                  Research Papers},
  booktitle    = {{NAACL-HLT}},
  pages        = {4599--4610},
  publisher    = {Association for Computational Linguistics},
  year         = {2021}
}

@inproceedings{yang2018hotpotqa,
  author       = {Zhilin Yang and
                  Peng Qi and
                  Saizheng Zhang and
                  Yoshua Bengio and
                  William W. Cohen and
                  Ruslan Salakhutdinov and
                  Christopher D. Manning},
  title        = {HotpotQA: {A} Dataset for Diverse, Explainable Multi-hop Question
                  Answering},
  booktitle    = {{EMNLP}},
  pages        = {2369--2380},
  publisher    = {Association for Computational Linguistics},
  year         = {2018}
}

@inproceedings{castelli2020techqa,
  author       = {Vittorio Castelli and
                  Rishav Chakravarti and
                  Saswati Dana and
                  Anthony Ferritto and
                  Radu Florian and
                  Martin Franz and
                  Dinesh Garg and
                  Dinesh Khandelwal and
                  J. Scott McCarley and
                  Mike McCawley and
                  Mohamed Nasr and
                  Lin Pan and
                  Cezar Pendus and
                  John F. Pitrelli and
                  Saurabh Pujar and
                  Salim Roukos and
                  Andrzej Sakrajda and
                  Avirup Sil and
                  Rosario Uceda{-}Sosa and
                  Todd Ward and
                  Rong Zhang},
  title        = {The TechQA Dataset},
  booktitle    = {{ACL}},
  pages        = {1269--1278},
  publisher    = {Association for Computational Linguistics},
  year         = {2020}
}

@misc{benchmarkheaven2026jevbench,
  title = {{J}ev{B}ench: a benchmark for {AI} decision models (v1.6.1)},
  author = {{Benchmark Heaven}},
  year = {2026},
  howpublished = {\url{https://github.com/fstandhartinger/jevbench}}
}

@misc{together2026tev1,
  title = {tev1-4{B}-experimental: an open-weight {J}ev-inspired decision model},
  author = {{Together AI}},
  year = {2026},
  howpublished = {\url{https://github.com/togethercomputer/tev1}}
}

@article{gao2023ragsurvey,
  author       = {Yunfan Gao and
                  Yun Xiong and
                  Xinyu Gao and
                  Kangxiang Jia and
                  Jinliu Pan and
                  Yuxi Bi and
                  Yi Dai and
                  Jiawei Sun and
                  Qianyu Guo and
                  Meng Wang and
                  Haofen Wang},
  title        = {Retrieval-Augmented Generation for Large Language Models: {A} Survey},
  journal      = {CoRR},
  volume       = {abs/2312.10997},
  year         = {2023}
}

@misc{baai2024rerankerv2m3,
  title = {bge-reranker-v2-m3},
  author = {{BAAI}},
  year = {2024},
  howpublished = {Hugging Face model card},
  url = {https://huggingface.co/BAAI/bge-reranker-v2-m3}
}

@inproceedings{yu2021convdr,
  author       = {Shi Yu and
                  Zhenghao Liu and
                  Chenyan Xiong and
                  Tao Feng and
                  Zhiyuan Liu},
  title        = {Few-Shot Conversational Dense Retrieval},
  booktitle    = {{SIGIR}},
  pages        = {829--838},
  publisher    = {{ACM}},
  year         = {2021}
}

@article{beyondbenchmarkscores2026,
  author       = {Filip J. Kucia and
                  Barbara M. Gawlik},
  title        = {Beyond Benchmark Scores: How Synthetic and Authentic Query Distributions
                  Diverge in {RAG} Evaluation},
  journal      = {CoRR},
  volume       = {abs/2609.14579},
  year         = {2026}
}

@inproceedings{guu2020realm,
  author       = {Kelvin Guu and
                  Kenton Lee and
                  Zora Tung and
                  Panupong Pasupat and
                  Ming{-}Wei Chang},
  title        = {Retrieval Augmented Language Model Pre-Training},
  booktitle    = {{ICML}},
  series       = {Proceedings of Machine Learning Research},
  volume       = {119},
  pages        = {3929--3938},
  publisher    = {{PMLR}},
  year         = {2020}
}

@inproceedings{izacard2021leveraging,
  author       = {Gautier Izacard and
                  Edouard Grave},
  title        = {Leveraging Passage Retrieval with Generative Models for Open Domain
                  Question Answering},
  booktitle    = {{EACL}},
  pages        = {874--880},
  publisher    = {Association for Computational Linguistics},
  year         = {2021}
}

@inproceedings{borgeaud2022improving,
  author       = {Sebastian Borgeaud and
                  Arthur Mensch and
                  Jordan Hoffmann and
                  Trevor Cai and
                  Eliza Rutherford and
                  Katie Millican and
                  George van den Driessche and
                  Jean{-}Baptiste Lespiau and
                  Bogdan Damoc and
                  Aidan Clark and
                  Diego de Las Casas and
                  Aurelia Guy and
                  Jacob Menick and
                  Roman Ring and
                  Tom Hennigan and
                  Saffron Huang and
                  Loren Maggiore and
                  Chris Jones and
                  Albin Cassirer and
                  Andy Brock and
                  Michela Paganini and
                  Geoffrey Irving and
                  Oriol Vinyals and
                  Simon Osindero and
                  Karen Simonyan and
                  Jack W. Rae and
                  Erich Elsen and
                  Laurent Sifre},
  title        = {Improving Language Models by Retrieving from Trillions of Tokens},
  booktitle    = {{ICML}},
  series       = {Proceedings of Machine Learning Research},
  volume       = {162},
  pages        = {2206--2240},
  publisher    = {{PMLR}},
  year         = {2022}
}

@article{gao2023retrieval,
  author       = {Yunfan Gao and
                  Yun Xiong and
                  Xinyu Gao and
                  Kangxiang Jia and
                  Jinliu Pan and
                  Yuxi Bi and
                  Yi Dai and
                  Jiawei Sun and
                  Qianyu Guo and
                  Meng Wang and
                  Haofen Wang},
  title        = {Retrieval-Augmented Generation for Large Language Models: {A} Survey},
  journal      = {CoRR},
  volume       = {abs/2312.10997},
  year         = {2023}
}

@inproceedings{kandpal2023large,
  author       = {Nikhil Kandpal and
                  Haikang Deng and
                  Adam Roberts and
                  Eric Wallace and
                  Colin Raffel},
  title        = {Large Language Models Struggle to Learn Long-Tail Knowledge},
  booktitle    = {{ICML}},
  series       = {Proceedings of Machine Learning Research},
  volume       = {202},
  pages        = {15696--15707},
  publisher    = {{PMLR}},
  year         = {2023}
}

@inproceedings{vu2024freshllms,
  author       = {Tu Vu and
                  Mohit Iyyer and
                  Xuezhi Wang and
                  Noah Constant and
                  Jerry W. Wei and
                  Jason Wei and
                  Chris Tar and
                  Yun{-}Hsuan Sung and
                  Denny Zhou and
                  Quoc V. Le and
                  Thang Luong},
  title        = {FreshLLMs: Refreshing Large Language Models with Search Engine
                  Augmentation},
  booktitle    = {{ACL} (Findings)},
  series       = {Findings of {ACL}},
  volume       = {ACL 2024},
  pages        = {13697--13720},
  publisher    = {Association for Computational Linguistics},
  year         = {2024}
}

@inproceedings{shuster2021retrieval,
  author       = {Kurt Shuster and
                  Spencer Poff and
                  Moya Chen and
                  Douwe Kiela and
                  Jason Weston},
  title        = {Retrieval Augmentation Reduces Hallucination in Conversation},
  booktitle    = {{EMNLP} (Findings)},
  series       = {Findings of {ACL}},
  volume       = {EMNLP 2021},
  pages        = {3784--3803},
  publisher    = {Association for Computational Linguistics},
  year         = {2021}
}

@article{menick2022teaching,
  author       = {Jacob Menick and
                  Maja Trebacz and
                  Vladimir Mikulik and
                  John Aslanides and
                  H. Francis Song and
                  Martin J. Chadwick and
                  Mia Glaese and
                  Susannah Young and
                  Lucy Campbell{-}Gillingham and
                  Geoffrey Irving and
                  Nat McAleese},
  title        = {Teaching language models to support answers with verified quotes},
  journal      = {CoRR},
  volume       = {abs/2203.11147},
  year         = {2022}
}

@inproceedings{gao2023enabling,
  author       = {Tianyu Gao and
                  Howard Yen and
                  Jiatong Yu and
                  Danqi Chen},
  title        = {Enabling Large Language Models to Generate Text with Citations},
  booktitle    = {{EMNLP}},
  pages        = {6465--6488},
  publisher    = {Association for Computational Linguistics},
  year         = {2023}
}

@article{openai2026gpt5,
  author       = {OpenAI},
  title        = {OpenAI {GPT-5} System Card},
  journal      = {CoRR},
  volume       = {abs/2601.03267},
  year         = {2026}
}

@inproceedings{brown2020language,
  author       = {Tom B. Brown and
                  Benjamin Mann and
                  Nick Ryder and
                  Melanie Subbiah and
                  Jared Kaplan and
                  Prafulla Dhariwal and
                  Arvind Neelakantan and
                  Pranav Shyam and
                  Girish Sastry and
                  Amanda Askell and
                  Sandhini Agarwal and
                  Ariel Herbert{-}Voss and
                  Gretchen Krueger and
                  Tom Henighan and
                  Rewon Child and
                  Aditya Ramesh and
                  Daniel M. Ziegler and
                  Jeffrey Wu and
                  Clemens Winter and
                  Christopher Hesse and
                  Mark Chen and
                  Eric Sigler and
                  Mateusz Litwin and
                  Scott Gray and
                  Benjamin Chess and
                  Jack Clark and
                  Christopher Berner and
                  Sam McCandlish and
                  Alec Radford and
                  Ilya Sutskever and
                  Dario Amodei},
  title        = {Language Models are Few-Shot Learners},
  booktitle    = {NeurIPS},
  year         = {2020}
}

@article{openai2023gpt4,
  author       = {OpenAI},
  title        = {{GPT-4} Technical Report},
  journal      = {CoRR},
  volume       = {abs/2303.08774},
  year         = {2023}
}

@article{grattafiori2024llama,
  author       = {Llama Team},
  title        = {The Llama 3 Herd of Models},
  journal      = {CoRR},
  volume       = {abs/2407.21783},
  year         = {2024}
}

@article{izacard2022unsupervised,
  author       = {Gautier Izacard and
                  Mathilde Caron and
                  Lucas Hosseini and
                  Sebastian Riedel and
                  Piotr Bojanowski and
                  Armand Joulin and
                  Edouard Grave},
  title        = {Unsupervised Dense Information Retrieval with Contrastive Learning},
  journal      = {Trans. Mach. Learn. Res.},
  volume       = {2022},
  year         = {2022}
}

@article{wang2022text,
  author       = {Liang Wang and
                  Nan Yang and
                  Xiaolong Huang and
                  Binxing Jiao and
                  Linjun Yang and
                  Daxin Jiang and
                  Rangan Majumder and
                  Furu Wei},
  title        = {Text Embeddings by Weakly-Supervised Contrastive Pre-training},
  journal      = {CoRR},
  volume       = {abs/2212.03533},
  year         = {2022}
}

@inproceedings{barnett2024seven,
  author       = {Scott Barnett and
                  Stefanus Kurniawan and
                  Srikanth Thudumu and
                  Zach Brannelly and
                  Mohamed Abdelrazek},
  title        = {Seven Failure Points When Engineering a Retrieval Augmented Generation
                  System},
  booktitle    = {{CAIN}},
  pages        = {194--199},
  publisher    = {{ACM}},
  year         = {2024}
}

@inproceedings{salemi2024evaluating,
  author       = {Alireza Salemi and
                  Hamed Zamani},
  title        = {Evaluating Retrieval Quality in Retrieval-Augmented Generation},
  booktitle    = {{SIGIR}},
  pages        = {2395--2400},
  publisher    = {{ACM}},
  year         = {2024}
}

@article{nakano2021webgpt,
  author       = {Reiichiro Nakano and
                  Jacob Hilton and
                  Suchir Balaji and
                  Jeff Wu and
                  Long Ouyang and
                  Christina Kim and
                  Christopher Hesse and
                  Shantanu Jain and
                  Vineet Kosaraju and
                  William Saunders and
                  Xu Jiang and
                  Karl Cobbe and
                  Tyna Eloundou and
                  Gretchen Krueger and
                  Kevin Button and
                  Matthew Knight and
                  Benjamin Chess and
                  John Schulman},
  title        = {WebGPT: Browser-assisted question-answering with human feedback},
  journal      = {CoRR},
  volume       = {abs/2112.09332},
  year         = {2021}
}

@inproceedings{yao2023react,
  author       = {Shunyu Yao and
                  Jeffrey Zhao and
                  Dian Yu and
                  Nan Du and
                  Izhak Shafran and
                  Karthik R. Narasimhan and
                  Yuan Cao},
  title        = {ReAct: Synergizing Reasoning and Acting in Language Models},
  booktitle    = {{ICLR}},
  publisher    = {OpenReview.net},
  year         = {2023}
}

@inproceedings{trivedi2023interleaving,
  author       = {Harsh Trivedi and
                  Niranjan Balasubramanian and
                  Tushar Khot and
                  Ashish Sabharwal},
  title        = {Interleaving Retrieval with Chain-of-Thought Reasoning for
                  Knowledge-Intensive Multi-Step Questions},
  booktitle    = {{ACL} {(1)}},
  pages        = {10014--10037},
  publisher    = {Association for Computational Linguistics},
  year         = {2023}
}

@article{chen2023walking,
  author       = {Howard Chen and
                  Ramakanth Pasunuru and
                  Jason Weston and
                  Asli Celikyilmaz},
  title        = {Walking Down the Memory Maze: Beyond Context Limit through Interactive
                  Reading},
  journal      = {CoRR},
  volume       = {abs/2310.05029},
  year         = {2023}
}

@article{jin2025searchr1,
  author       = {Bowen Jin and
                  Hansi Zeng and
                  Zhenrui Yue and
                  Dong Wang and
                  Hamed Zamani and
                  Jiawei Han},
  title        = {Search-R1: Training LLMs to Reason and Leverage Search Engines with
                  Reinforcement Learning},
  journal      = {CoRR},
  volume       = {abs/2503.09516},
  year         = {2025}
}

@article{nogueira2019passage,
  author       = {Rodrigo Nogueira and
                  Kyunghyun Cho},
  title        = {Passage Re-ranking with {BERT}},
  journal      = {CoRR},
  volume       = {abs/1901.04085},
  year         = {2019}
}

@inproceedings{nogueira2020document,
  author       = {Rodrigo Nogueira and
                  Zhiying Jiang and
                  Ronak Pradeep and
                  Jimmy Lin},
  title        = {Document Ranking with a Pretrained Sequence-to-Sequence Model},
  booktitle    = {{EMNLP} (Findings)},
  series       = {Findings of {ACL}},
  volume       = {EMNLP 2020},
  pages        = {708--718},
  publisher    = {Association for Computational Linguistics},
  year         = {2020}
}

@inproceedings{sun2023chatgpt,
  author       = {Weiwei Sun and
                  Lingyong Yan and
                  Xinyu Ma and
                  Shuaiqiang Wang and
                  Pengjie Ren and
                  Zhumin Chen and
                  Dawei Yin and
                  Zhaochun Ren},
  title        = {Is ChatGPT Good at Search? Investigating Large Language Models as
                  Re-Ranking Agents},
  booktitle    = {{EMNLP}},
  pages        = {14918--14937},
  publisher    = {Association for Computational Linguistics},
  year         = {2023}
}

@inproceedings{chen2025attention,
  author       = {Shijie Chen and
                  Bernal Jimenez Gutierrez and
                  Yu Su},
  title        = {Attention in Large Language Models Yields Efficient Zero-Shot Re-Rankers},
  booktitle    = {{ICLR}},
  publisher    = {OpenReview.net},
  year         = {2025}
}

@inproceedings{gupta2025blockrank,
  author       = {Nilesh Gupta and
                  Chong You and
                  Srinadh Bhojanapalli and
                  Sanjiv Kumar and
                  Inderjit S. Dhillon and
                  Felix X. Yu},
  title        = {Scalable In-context Ranking with Generative Models},
  booktitle    = {NeurIPS},
  year         = {2025}
}

@article{wang2025jinav3,
  author       = {Feng Wang and
                  Yuqing Li and
                  Han Xiao},
  title        = {jina-reranker-v3: Last but Not Late Interaction for Document Reranking},
  journal      = {CoRR},
  volume       = {abs/2509.25085},
  year         = {2025}
}

@inproceedings{sachan2022improving,
  author       = {Devendra Singh Sachan and
                  Mike Lewis and
                  Mandar Joshi and
                  Armen Aghajanyan and
                  Wen{-}tau Yih and
                  Joelle Pineau and
                  Luke Zettlemoyer},
  title        = {Improving Passage Retrieval with Zero-Shot Question Generation},
  booktitle    = {{EMNLP}},
  pages        = {3781--3797},
  publisher    = {Association for Computational Linguistics},
  year         = {2022}
}

@article{pirolli1999information,
  title   = {Information Foraging},
  author  = {Pirolli, Peter and Card, Stuart},
  journal = {Psychological Review},
  volume  = {106},
  number  = {4},
  pages   = {643--675},
  year    = {1999}
}

@inproceedings{duggan2011skim,
  author       = {Geoffrey B. Duggan and
                  Stephen J. Payne},
  title        = {Skim reading by satisficing: evidence from eye tracking},
  booktitle    = {{CHI}},
  pages        = {1141--1150},
  publisher    = {{ACM}},
  year         = {2011}
}

@article{edge2024local,
  author       = {Darren Edge and
                  Ha Trinh and
                  Newman Cheng and
                  Joshua Bradley and
                  Alex Chao and
                  Apurva Mody and
                  Steven Truitt and
                  Jonathan Larson},
  title        = {From Local to Global: {A} Graph {RAG} Approach to Query-Focused
                  Summarization},
  journal      = {CoRR},
  volume       = {abs/2404.16130},
  year         = {2024}
}

@article{kwiatkowski2019natural,
  author       = {Tom Kwiatkowski and
                  Jennimaria Palomaki and
                  Olivia Redfield and
                  Michael Collins and
                  Ankur P. Parikh and
                  Chris Alberti and
                  Danielle Epstein and
                  Illia Polosukhin and
                  Jacob Devlin and
                  Kenton Lee and
                  Kristina Toutanova and
                  Llion Jones and
                  Matthew Kelcey and
                  Ming{-}Wei Chang and
                  Andrew M. Dai and
                  Jakob Uszkoreit and
                  Quoc Le and
                  Slav Petrov},
  title        = {Natural Questions: a Benchmark for Question Answering Research},
  journal      = {Trans. Assoc. Comput. Linguistics},
  volume       = {7},
  pages        = {452--466},
  year         = {2019}
}

@article{adlakha2022topiocqa,
  author       = {Vaibhav Adlakha and
                  Shehzaad Dhuliawala and
                  Kaheer Suleman and
                  Harm de Vries and
                  Siva Reddy},
  title        = {TopiOCQA: Open-domain Conversational Question Answering with Topic
                  Switching},
  journal      = {Trans. Assoc. Comput. Linguistics},
  volume       = {10},
  pages        = {468--483},
  year         = {2022}
}

@inproceedings{anantha2021open,
  author       = {Raviteja Anantha and
                  Svitlana Vakulenko and
                  Zhucheng Tu and
                  Shayne Longpre and
                  Stephen Pulman and
                  Srinivas Chappidi},
  title        = {Open-Domain Question Answering Goes Conversational via Question
                  Rewriting},
  booktitle    = {{NAACL-HLT}},
  pages        = {520--534},
  publisher    = {Association for Computational Linguistics},
  year         = {2021}
}

@book{manning2008xml,
  author       = {Christopher D. Manning and
                  Prabhakar Raghavan and
                  Hinrich Sch{\"{u}}tze},
  title        = {Introduction to information retrieval},
  publisher    = {Cambridge University Press},
  year         = {2008}
}

@inproceedings{liu2021dense,
  author       = {Ye Liu and
                  Kazuma Hashimoto and
                  Yingbo Zhou and
                  Semih Yavuz and
                  Caiming Xiong and
                  Philip S. Yu},
  title        = {Dense Hierarchical Retrieval for Open-domain Question Answering},
  booktitle    = {{EMNLP} (Findings)},
  series       = {Findings of {ACL}},
  volume       = {EMNLP 2021},
  pages        = {188--200},
  publisher    = {Association for Computational Linguistics},
  year         = {2021}
}

@inproceedings{jeong2024adaptiverag,
  author       = {Soyeong Jeong and
                  Jinheon Baek and
                  Sukmin Cho and
                  Sung Ju Hwang and
                  Jong C. Park},
  title        = {Adaptive-RAG: Learning to Adapt Retrieval-Augmented Large Language
                  Models through Question Complexity},
  booktitle    = {{NAACL-HLT}},
  pages        = {7036--7050},
  publisher    = {Association for Computational Linguistics},
  year         = {2024}
}

@inproceedings{yang2026searchagentx,
  author       = {Tiannuo Yang and
                  Zebin Yao and
                  Bowen Jin and
                  Lixiao Cui and
                  Yusen Li and
                  Gang Wang and
                  Xiaoguang Liu and
                  Willie Neiswanger},
  title        = {Demystifying and Enhancing the Efficiency of Large Language Model Based
                  Search Agents},
  booktitle    = {{ICLR}},
  publisher    = {proceedings.iclr.cc},
  year         = {2026}
}

@inproceedings{cutting1992scatter,
  author       = {Douglas R. Cutting and
                  Jan O. Pedersen and
                  David R. Karger and
                  John W. Tukey},
  title        = {Scatter/Gather: {A} Cluster-based Approach to Browsing Large Document
                  Collections},
  booktitle    = {{SIGIR}},
  pages        = {318--329},
  publisher    = {{ACM}},
  year         = {1992}
}

@article{gupta2026lattice,
  author       = {Nilesh Gupta and
                  Wei{-}Cheng Chang and
                  Ngot Bui and
                  Cho{-}Jui Hsieh and
                  Inderjit S. Dhillon},
  title        = {LLM-guided Hierarchical Search for End-to-end Reasoning Intensive
                  Retrieval},
  journal      = {Trans. Mach. Learn. Res.},
  volume       = {2026},
  year         = {2026}
}

@article{zhang2026structures,
  author       = {Yeqin Zhang and
                  Haomin Fu and
                  Xujie Zhang and
                  Cam{-}Tu Nguyen},
  title        = {Structures Facilitate Retrieve, Rerank, and Generate},
  journal      = {CoRR},
  volume       = {abs/2606.03247},
  year         = {2026}
}

@inproceedings{nister2006vocabtree,
  author       = {David Nist{\'{e}}r and
                  Henrik Stew{\'{e}}nius},
  title        = {Scalable Recognition with a Vocabulary Tree},
  booktitle    = {{CVPR} {(2)}},
  pages        = {2161--2168},
  publisher    = {{IEEE} Computer Society},
  year         = {2006}
}

@inproceedings{monath2021scc,
  author       = {Nicholas Monath and
                  Kumar Avinava Dubey and
                  Guru Guruganesh and
                  Manzil Zaheer and
                  Amr Ahmed and
                  Andrew McCallum and
                  G{\"{o}}khan Mergen and
                  Marc Najork and
                  Mert Terzihan and
                  Bryon Tjanaka and
                  Yuan Wang and
                  Yuchen Wu},
  title        = {Scalable Hierarchical Agglomerative Clustering},
  booktitle    = {{KDD}},
  pages        = {1245--1255},
  publisher    = {{ACM}},
  year         = {2021}
}

@inproceedings{bateni2017affinity,
  author       = {MohammadHossein Bateni and
                  Soheil Behnezhad and
                  Mahsa Derakhshan and
                  MohammadTaghi Hajiaghayi and
                  Raimondas Kiveris and
                  Silvio Lattanzi and
                  Vahab S. Mirrokni},
  title        = {Affinity Clustering: Hierarchical Clustering at Scale},
  booktitle    = {{NIPS}},
  pages        = {6864--6874},
  year         = {2017}
}

@article{dhulipala2023terahac,
  author       = {Laxman Dhulipala and
                  Jakub Lacki and
                  Jason Lee and
                  Vahab Mirrokni},
  title        = {TeraHAC: Hierarchical Agglomerative Clustering of Trillion-Edge Graphs},
  journal      = {Proc. {ACM} Manag. Data},
  volume       = {1},
  number       = {3},
  pages        = {221:1--221:27},
  year         = {2023}
}

@article{yang2025qwen3technical,
  author       = {Qwen Team},
  title        = {Qwen3 Technical Report},
  journal      = {CoRR},
  volume       = {abs/2505.09388},
  year         = {2025}
}
\bibliographystyle{icml2026}

\newpage
\appendix
\crefalias{section}{appendix}
\crefalias{subsection}{appendix}
\onecolumn
\raggedbottom  % the appendix's [H] tables otherwise leave stretched gaps
\section{Dataset and Corpus Details}
\label{app:datasets}

\subsection{MultiDoc2Dial}
\label{app:datasets:md2d}

MultiDoc2Dial \citep{feng2021multidoc2dial} contains
goal-oriented dialogues grounded in 488 documents from
four US government help centers. We build a
help-center--page--passage hierarchy from the dataset's
metadata. Help centers are given by the domain field, and
documents sharing a title after removing the \#1/\#2 part
suffix form one page, giving 413 pages. We use the
benchmark's official structure-based segmentation into
4,110 passages.

Passages are represented by their full text and their page
and parent-heading path, pages by their titles and section
headings, and the four help centers by one-line descriptions
written manually from their names. We evaluate on the
official validation split (661 dialogues, 4,201 turns). Each
turn has one gold passage containing the grounding span of
the agent's response. The released test split provides no
usable evaluation examples.

\subsection{ConditionalQA}
\label{app:datasets:conditionalqa}

ConditionalQA \citep{sun2022conditionalqa} contains
scenario-based questions about UK government guidance whose
answers may depend on several conditions and paragraphs. Its
corpus contains 652 guides. We recover their structure from
the HTML headings (\texttt{h1}, \texttt{h2}) and extract
58,483 paragraph-level units from \texttt{p}, \texttt{li} and
\texttt{tr} elements. Pages are represented by their titles
and top-level part headings, and paragraphs are grouped under
their heading paths. Retrieval searches the full corpus rather
than assuming the relevant guide is known. We evaluate on the
285 questions of the labeled development split, 271 of which
have annotated evidence (about four paragraphs on average).

\subsection{MTRAG--IBM Cloud}
\label{app:datasets:mtrag}

We use the IBM Cloud subset of MTRAG \citep{katsis2025mtrag},
with 8,578 documentation pages and 72,442 passages, keeping the
benchmark's document files and passage chunks. Pages are grouped
into 172 products by the first path segment after
\texttt{/docs/} in their URLs. Because the corpus does not
preserve HTML markup, we recover section headings from the text
layout (short lines separated by blank lines). Pages are
represented by their titles and recovered headings, and passages
by their text and heading. Product descriptions are built
automatically from the product name, the opening of its landing
page and three page titles.

The held-out evaluation set contains 26 human-written
conversations with 205 turns, 188 of which have annotated
evidence (one to seven gold passages).

\paragraph{Expanded relevance judgments.}
MTRAG's gold passages were selected from the annotators' ELSER retrieval
sessions, so relevant passages that ELSER did not surface cannot be gold.
We therefore add LLM-judged relevant passages to the gold passages. For
each turn, we pool the top five passages of all 18 systems evaluated on
MTRAG, including released ELSER and the last-turn and rewritten-query
baselines, together with the gold passages, representing identical texts
once. GPT-5 receives the conversation, the reference reply and the
shuffled candidates, without knowing which system retrieved them, and
grades each passage as needed for the answer (2), related but
insufficient (1) or unrelated (0). Grade-2 passages are added to the
relevance set, and all original gold passages are kept. GPT-5 grades
87.9\% of the gold passages as needed. The expanded labels reduce the
dependence on ELSER. \cref{tab:mtrag-judged} compares all systems.

\subsection{Natural Questions}
\label{app:datasets:nq}

We use the DAPR version \citep{wang2024dapr} of Natural
Questions \citep{kwiatkowski2019natural}, whose corpus splits
108,626 Wikipedia pages into 2,682,017 paragraphs. We evaluate
on a fixed random sample of 500 of the 3,610 questions in DAPR's
test split, which is drawn from NQ's development set. A
question's gold paragraphs (one to three) lie on a single page.
Since Wikipedia has no structure above the page, we cluster the
pages into 1,000 groups with $k$-means over e5-base-v2
embeddings \citep{wang2022text} of each page's title and first
sentence. Groups hold
15 to 350 pages (median 106). Each group is described by one
sentence that Qwen3-8B writes from its page titles
(\cref{app:retrieval:nq}). Pages are represented by their titles
and paragraphs by their text, grouped under their page title.

\subsection{Design Sets and Configuration Selection}
\label{app:datasets:design}

Retrieval configurations (candidate representations, routing
rules and retention budgets) are selected on design sets that
are disjoint from the evaluation sets. Selection does not update
the parameters of \jev{}. MultiDoc2Dial uses samples of its
training split and ConditionalQA 300 training questions. MTRAG has no training split, so
we use 50 synthetic IBM Cloud conversations (292 turns) and 131
IBM Cloud turns from MTRAG-UN \citep{rosenthal2026mtragun},
whose conversations are disjoint from the evaluation set.
 
\section{Retrieval Configuration Details}
\label{app:retrieval}

\subsection{Candidate Representations and Questions}
\label{app:retrieval:prompts}

At each stage, the judge receives a context containing the
query and the numbered candidate descriptions, followed by one
yes/no question per candidate. Candidates are listed in document
order and grouped under their parent pages where applicable. Each
stage builds a new context; probabilities from earlier stages are
not included. Conversational queries contain the six most recent
dialogue turns.

For conversational retrieval, the page and passage questions are
``Is the information the agent needs for its next reply on
Page~$n$?'' and ``Should the agent's next reply be based on
Passage~$n$?'' For ConditionalQA and Natural Questions, they are
``Is the information needed to answer the user's question on
Page~$n$?'' and ``Does Paragraph~$n$ contain information needed
to answer the user's question?''

\subsection{MultiDoc2Dial and the Conversational Notepad}
\label{app:retrieval:md2d}

The routing context describes the four help centers, with one
question per center. We keep the highest-scoring center, the
runner-up if the top probability is below 0.8, and the center of
the most recently retrieved page. The page stage scores the pages
of the kept centers and keeps ten. All passages of these pages,
plus those of the two most recent notepad pages, are scored in one
shared context.

The notepad, used on MultiDoc2Dial and MTRAG, lists distinct page
identifiers by most recent retrieval. After each turn, the page of
the top-ranked passage moves to the front. The page-stage context
lists the notepad pages. The notepad is built only from the system's
own retrieval decisions, never from gold evidence, and previously
retrieved passages are neither marked nor excluded.

\subsection{ConditionalQA}
\label{app:retrieval:conditionalqa}

The page stage scores all 652 guides in one request and keeps three,
whose paragraphs are scored together under their heading paths.

\subsection{MTRAG}
\label{app:retrieval:mtrag}

The main configuration scores pages directly instead of routing
through products. In a first pass, all 8,578 page titles are split
into parallel requests under the input-size limit, keeping each
product's pages together, and the top 50 pages by pooled score are
kept. Because these scores come from different contexts, a
second pass rescores the 50 together with their product names and
headings, and keeps ten.

Pages with more than 60 passages pass through an extra stage that
scores their headings and keeps the top eight. At most 60 passages
of these headings, in document order, continue to the passage stage.

\subsection{Natural Questions}
\label{app:retrieval:nq}

Each cluster is described by Qwen3-8B, given its page titles (most
typical first, at most 300) and the prompt: ``Write one sentence (at
most 30 words) describing what the pages in this group are about.
Name the main kinds of subjects it covers, including the less typical
ones, so that a reader can tell whether a given Wikipedia page would
belong to this group.'' The judge scores the 1,000 cluster
descriptions in eight parallel requests, asking for each whether
``the page with the information needed to answer the user's question
[is] likely in Topic~$n$'', and keeps 150 clusters. Their page
titles, about 17,500 per question, are scored in parallel requests
of at most 20k tokens. The paragraph stage scores all paragraphs of
the top three pages, the first ten of pages ranked 4--10 and the
first three of pages ranked 11--30 in one shared context, about 230
paragraphs per question. Each paragraph's probability is multiplied
by its page's.

\subsection{Multiple-Choice Ranking}
\label{app:retrieval:choice}
The multiple-choice question follows binary scoring. Its answer options
are the top 20 candidates by $P(\mathrm{yes})$ (top five on Natural
Questions), and its context lists only these candidates, in reading
order and formatted as in binary scoring. On MultiDoc2Dial,
ConditionalQA and Natural Questions, the choice probabilities reorder
the shortlist, and all other candidates keep their binary order. On
MTRAG, only the highest-probability choice moves to rank one. Compared
with a choice request over the full binary context, the shorter context
changes retrieval by at most one point on every dataset and shortens
the MTRAG choice request from 0.30--0.37 to 0.14 seconds. Retrieval
without the choice step is evaluated in \cref{sec:ablation:choice}.
 
\section{Full Benchmark Results}\label{app:full-results}
The tables below extend \cref{tab:retrieval-summary} with secondary
metrics, published reference results, task-trained systems and the
full MTRAG comparison; cost
and latency are in \cref{tab:latency} and \cref{app:timing}. Rows named as in \cref{tab:retrieval-summary} are the same
configurations. A dash denotes a result or
measurement that is unavailable, never a zero. Open-weight judge results are in
\cref{tab:judge-full}.
\begin{table}[H]
\centering
\small
\setlength{\tabcolsep}{4pt}
\caption{MultiDoc2Dial validation (4{,}201 turns), passage recall by
passage ID (\%). Dense: Qwen3-Embedding-8B; BGE: bge-reranker-v2-m3.
$^\dagger$Published results, not rerun.}
\label{tab:md2d-extended}
\begin{tabular}{@{}lrrr@{}}
\toprule
Method & R@1 & R@5 & R@10 \\
\midrule
\multicolumn{4}{@{}l}{\textbf{Conventional retrieval pipelines}} \\
BM25 \citep{feng2021multidoc2dial}$^\dagger$ & 19.6 & 41.9 & 50.8 \\
BM25 (ours) & 20.6 & 46.8 & 56.4 \\
SPLADE++ & 24.4 & -- & 58.8 \\
Dense & 27.9 & 60.4 & 70.9 \\
Dense top-100 + BGE & 22.8 & 52.8 & 63.9 \\
Dense top-100 + Qwen3-Reranker & 33.6 & 71.8 & 81.1 \\
\midrule
\multicolumn{4}{@{}l}{\textbf{LLM-based listwise reranking}} \\
Dense top-20 + GPT-5$^s$ & 47.5 & 73.7 & 78.6 \\
\midrule
\multicolumn{4}{@{}l}{\textbf{Task-trained retrieval}} \\
DPR fine-tuned \citep{feng2021multidoc2dial}$^\dagger$ & 49.0 & 72.3 & 80.0 \\
DPR + cross-encoder \citep{sfre2g2026}$^\dagger$ & 55.0 & 78.3 & 83.2 \\
\textbf{\method{} (8B trained), 3 pages} & 60.7 & 82.7 & 85.8 \\
\textbf{\method{} (8B trained), 10 pages} & 60.8 & 86.1 & 90.0 \\
\midrule
\multicolumn{4}{@{}l}{\textbf{Decision-based retrieval (ours)}} \\
\textbf{\method{} (\jev{})} & 47.4 & 76.2 & 83.6 \\
\textbf{\method{} (\jev{}) + dense candidates} & 48.5 & 77.6 & 84.7 \\
\bottomrule
\end{tabular}
\vspace{2pt}
\begin{minipage}{0.62\textwidth}\footnotesize
$^s$Seeded sample of 1,000 turns, on which the main row reaches
49.9 / 77.4 / 83.9. Cost and latency are in \cref{tab:latency,tab:cost}.
\end{minipage}
\end{table}

\begin{table}[H]
\centering
\small
\setlength{\tabcolsep}{3pt}
\caption{ConditionalQA development set, retrieval from all 652 pages
(\%). Page metrics use 285 questions and paragraph metrics the 271
questions with evidence. The paragraph column shows Hit@1 above
R@1 / R@5 / R@10. nDCG@10 uses DAPR's relevance judgments
\citep{wang2024dapr}. Dense and BGE as in \cref{tab:md2d-extended}.}
\label{tab:cqa-extended}
\begin{tabular*}{\textwidth}{@{\extracolsep{\fill}}lccc@{}}
\toprule
Method & \shortstack{Page\\Hit@1 / 3 / 10} & \shortstack{Paragraph\\Hit@1; R@1 / 5 / 10} & \shortstack{nDCG@10\\(DAPR)} \\
\midrule
\multicolumn{4}{@{}l}{\textbf{Conventional retrieval pipelines}} \\
BM25 + headings & 51.6 / 74.0 / 89.1 & \shortstack{24.7\\8.9 / 24.6 / 35.0} & 27.5 \\
SPLADE-v3 (DistilBERT) & 62.5 / 80.4 / 94.0 & \shortstack{26.9\\12.1 / 26.9 / 34.1} & 28.1 \\
SPLADE++ & 61.4 / 81.4 / 91.9 & \shortstack{28.4\\12.1 / 28.3 / 38.3} & 31.2 \\
Dense & 74.7 / 88.4 / 96.1 & \shortstack{47.2\\21.0 / 43.2 / 54.0} & 47.6 \\
Dense top-100 + BGE & 72.3 / 88.8 / 96.5 & \shortstack{53.1\\22.1 / 48.9 / 60.6} & 53.5 \\
Dense top-100 + Qwen3-Reranker & 68.8 / 78.9 / 95.4 & \shortstack{64.6\\29.3 / 54.8 / 67.2} & 62.2 \\
\midrule
\multicolumn{4}{@{}l}{\textbf{LLM-based listwise reranking}} \\
Dense top-20 + GPT-5 & -- & \shortstack{73.4\\30.4 / 57.6 / 64.0} & 63.4 \\
\midrule
\multicolumn{4}{@{}l}{\textbf{Task-trained retrieval}} \\
\textbf{\method{} (8B trained), 3 pages} & -- & \shortstack{75.3\\28.4 / 66.7 / 78.8} & 75.7 \\
\textbf{\method{} (8B trained), 5 pages} & -- & \shortstack{77.1\\29.9 / 67.7 / 80.2} & 77.3 \\
\midrule
\multicolumn{4}{@{}l}{\textbf{Decision-based retrieval (ours)}} \\
\textbf{\method{} (\jev{})} & -- & \shortstack{69.7\\27.3 / 59.4 / 71.9} & 67.5 \\
\textbf{\method{} (\jev{}), 5 pages} & 74.0 / 90.2 / 97.5 & \shortstack{71.6\\28.3 / 61.3 / 73.2} & -- \\
\textbf{\method{} (\jev{}) + dense candidates} & -- & \shortstack{73.4\\29.9 / 63.2 / 76.3} & 71.2 \\
\bottomrule
\end{tabular*}
\vspace{2pt}
\begin{minipage}{\textwidth}\footnotesize
The top-100 rerankers rerank the dense top 20 pages for the page column.
Cost and latency are in \cref{tab:latency,tab:cost}.
\end{minipage}
\end{table}

\paragraph{MTRAG.} \Cref{tab:mtrag-judged} compares all systems
against the gold and judged-relevant passages of
\cref{app:datasets:mtrag}. \method{}'s R@5 exceeds released ELSER by
7.6 points (95\% interval $[2.2, 13.1]$), last-turn SPLADE-v3 by 6.8
$[1.9, 12.3]$ and six-turn Qwen3-Reranker by 5.8 $[1.9, 9.6]$; it is
also ahead of last-turn Qwen3-Reranker and GPT-5 reranking.

\begin{table}[H]
\centering\small
\setlength{\tabcolsep}{3pt}
\caption{MTRAG, 188 labeled turns, against gold and judged-relevant
passages (\%); identical texts count once. The remaining nine systems
(BM25, dense, SPLADE and BGE variants) trail \method{} by 10 to 20 R@5
points.}
\label{tab:mtrag-judged}
\begin{tabular}{@{}lrr@{}}
\toprule
System & R@5 & nDCG@5 \\
\midrule
\textbf{\method{} + dense candidates} & 36.6 & 65.0 \\
\textbf{\method{}} & 34.2 & 62.3 \\
Dense + Qwen3-Reranker, last turn & 32.5 & 56.5 \\
Dense top-20 + GPT-5 & 31.8 & 59.4 \\
Dense + Qwen3-Reranker, six turns & 28.4 & 50.9 \\
SPLADE++, last turn & 27.8 & 47.1 \\
SPLADE-v3, last turn & 27.4 & 47.8 \\
ELSER (released, rewritten query) & 26.6 & 46.4 \\
Dense, last turn & 26.0 & 46.5 \\
\bottomrule
\end{tabular}
\end{table}

\subsection{Retrieval-augmented generation}\label{app:rag}
The generator is Qwen3.5-27B (FP8 weights, thinking disabled, greedy decoding),
served with vLLM. Its input is a short task instruction, the passages numbered
in each system's rank order, and the query: the user's situation and question
on ConditionalQA, the question on Natural Questions, and the last six turns on
MTRAG. It answers ConditionalQA with ``Answer:'' followed by
yes, no, a short phrase or ``not answerable'' (optionally with conditions),
Natural Questions with a phrase of at most five words, and MTRAG with a reply
that says when the passages do not contain the answer. Exact match uses the standard answer normalization; the
Natural Questions references are the short answers of NQ-open, available for
334 of the 500 questions. The judge is the same model, shown the query, the
reference answer(s) and the generated answer, and asked whether the answer is
correct (for MTRAG, consistent with the reference response, and a refusal on
unanswerable turns). Part of the generations and verdicts were produced by the same model through
the OpenRouter API instead of our local deployment; on 200 prompts answered by
both, answer quality matched and the verdicts agreed on 96.5\%.

\section{Additional Analyses}\label{app:analyses}

\subsection{Shared context and conversational memory}\label{app:context}
The configurations in \cref{tab:context} use 297 turns from 50 complete
training dialogues, six-turn input, title-only page descriptions, ten
newly selected pages, no help-center routing and no choice step, so
their scores are not comparable to the main results. \emph{Separate}
scoring builds one context per page, containing the conversation, the
page title and its passages with their heading paths; \emph{shared}
scoring lists the passages of all kept pages in one context, grouped by
page. Both rank passages by $P(\mathrm{yes})$ for ``Should the agent's
next reply be based on Passage~$n$?''. \emph{Own notepad} is the page
memory of \cref{app:retrieval:md2d}, built by each configuration from
its own decisions. Paired R@1 differences, resampling whole dialogues,
are $+4.0$ $[0.7, 8.0]$ for adding the notepad and $+7.7$ $[3.3, 12.2]$
for shared over separate scoring with notepads.

\begin{table}[H]\centering\small
\caption{MultiDoc2Dial passage context and page memory on 297 design
turns (\%).}\label{tab:context}
\begin{tabular}{@{}llrrr@{}}\toprule
Passage context & Notepad & R@1 & R@5 & R@10\\\midrule
Separate & None & 32.7 & 54.5 & 63.6\\
Separate & Own & 36.7 & 61.3 & 70.4\\
Shared & Own & 44.4 & 69.7 & 76.8\\\bottomrule
\end{tabular}
\end{table}

\subsection{Remaining retrieval errors}\label{app:errors}
Given the gold page, passage scoring on MultiDoc2Dial design turns
reaches 62.7 R@1 and 94.3 R@5. On the ConditionalQA development set, the
three-page configuration without multiple-choice ranking misses the gold
page for 8.5\% of evidence-labeled questions, ranks a paragraph from
another page first for 9.6\%, and ranks another paragraph of the gold
page first for 13.7\%. Passage ranking thus remains a substantial source
of error even when navigation reaches the relevant page.

\subsection{Dense candidates and multiple-choice ranking}\label{app:laststage}
\emph{+ dense candidates} adds the top 20 dense-retrieved passages (top
10 on ConditionalQA) to the navigated candidates, removes duplicates,
and groups all candidates by page and section in reading order.
Multiple-choice ranking either reorders the top 20 (\emph{order}) or
moves only its pick to rank one (\emph{top1}). \Cref{tab:laststage-design}
reports the design-set comparisons, at three pages on MultiDoc2Dial and
ConditionalQA and ten pages on MTRAG-UN. The design sets are 1,832 turns
from 300 MultiDoc2Dial training dialogues, the 285 of the 300
ConditionalQA design questions that have paragraph evidence, and 86
labeled MTRAG-UN turns, one per conversation. $\Delta$ is the difference
in the primary metric to the configuration with neither component, with
an unadjusted paired 95\% interval. With larger page budgets (ten pages
on MultiDoc2Dial, five on ConditionalQA) the gains shrink: at most
$+0.6$ R@1 on MultiDoc2Dial, with every interval containing zero, and
at most $+4.2$ Hit@1 on ConditionalQA.

\begin{table}[H]\centering\small
\caption{Last-stage design comparisons (\%). Primary metric: R@1
(MultiDoc2Dial), Hit@1 (ConditionalQA), R@5 (MTRAG-UN).}
\label{tab:laststage-design}
\begin{tabular}{@{}lrrr@{}}\toprule
Configuration & Primary & R@10 & $\Delta$ primary [95\% CI]\\\midrule
\multicolumn{4}{@{}l}{\textit{MultiDoc2Dial}}\\
3 pages & 48.2 & 82.5 & --\\
3 pages + dense 20 & 50.1 & 86.1 & +1.9 [+0.9, +3.0]\\
3 pages + dense 20 + multiple-choice & 51.3 & 86.6 & +3.1 [+1.5, +4.7]\\
\addlinespace
\multicolumn{4}{@{}l}{\textit{ConditionalQA}}\\
3 pages & 71.2 & 74.1 & --\\
3 pages + dense 10 & 71.9 & 76.5 & +0.7 [$-$1.4, +2.8]\\
3 pages + dense 10 + multiple-choice & 76.5 & 78.0 & +5.3 [+1.1, +9.5]\\
\addlinespace
\multicolumn{4}{@{}l}{\textit{MTRAG-UN}}\\
10 pages & 52.3 & 56.4 & --\\
10 pages + dense 20 & 57.8 & 66.2 & +5.5 [+0.0, +11.4]\\
10 pages + dense 20 + multiple-choice (top1) & 57.5 & 66.2 & +5.2 [$-$0.3, +11.2]\\
\bottomrule
\end{tabular}
\end{table}

\section{Evaluation Sensitivity}\label{sec:results:labels}
We rescore the retrieval results under two alternative relevance
definitions.

\subsection{Identical passages under different IDs}\label{app:labels:text}
A retrieved passage can repeat the gold text under a different ID. We
group passages by their text after whitespace normalization (and
lowercasing for MTRAG), map retrieved and gold IDs to these groups, and
keep only the first occurrence of each group in a ranking, so duplicate
copies cannot create multiple hits.

Text matching raises MultiDoc2Dial R@1 to 55.8 for \method{} (57.0 with
dense candidates) and to 40.3 for dense retrieval with Qwen3-Reranker; the
ordering of systems is unchanged.

\subsection{ConditionalQA under DAPR relevance judgments}\label{app:labels:dapr}
DAPR provides alternative relevance judgments for ConditionalQA,
evaluated with nDCG@10 \citep{wang2024dapr} (last column of
\cref{tab:cqa-extended}). On the same 271 questions, \method{}'s nDCG@10
exceeds dense retrieval with Qwen3-Reranker by $+5.3$ points (95\%
interval $[1.3, 9.2]$) and GPT-5 by $+4.0$ $[0.3, 7.8]$, and its DAPR
R@10 exceeds GPT-5 by $+8.2$ $[4.0, 12.3]$. With dense candidates,
nDCG@10 rises to 71.2.

\section{Latency and Cost}
\label{app:timing}

\paragraph{\method{} timing.}
Timing runs call the live API without reading cached answers, process
one query at a time with up to eight concurrent requests within a
query, and run from an AMD EPYC 9745 client. The timer covers complete
retrieval, including retries and waiting, and starts after
initialization (corpus loading); no warm-up query is discarded, and
quantiles are computed from per-query totals. Each stage waits for the
preceding one; the choice request follows the binary request over the
top candidates only and adds 0.12 to 0.15 seconds. Rows with dense
candidates (H) load saved dense candidates and exclude dense encoding
and search, so they are not end-to-end latencies. MultiDoc2Dial timing
uses 48 turns from eight complete training dialogues, each
configuration with its own notepad; ConditionalQA uses 50 development
questions, MTRAG 24 turns from three human conversations, and Natural
Questions the first 100 of its 500 test questions. All rows were
measured from the same client, the MultiDoc2Dial rows in a later
session; small differences between live runs can reflect provider
variation. The main MTRAG configuration sends the first page pass as
requests of at most 10k tokens instead of 150k (about 40 parallel
requests instead of three); on the 188 held-out turns this leaves R@5
and nDCG@5 unchanged (paired differences $-0.7$ and $-0.2$ points,
intervals containing zero) and reduces median latency from 3.09 to 1.76
seconds.

\begin{table}[H]\centering\small
\caption{Online timing per query. $n$: timed queries; W: complete
retrieval after initialization; H: retrieval and scoring with saved
dense candidates; R: GPT-5 listwise reranking of the saved dense top 20,
API only.}\label{tab:cost}
\begin{tabular}{@{}llcrrr@{}}\toprule
Dataset & Configuration & Scope & $n$ & p50/p95 (s) & API ¢/query\\\midrule
MultiDoc2Dial & \method{} (\jev{}) & W & 48 & 0.68/0.91 & 0.22\\
 & \quad + dense candidates & H & 48 & 0.70/1.16 & 0.23\\
 & Dense top-20 + GPT-5 & R & 1,000 & 11.8/23.3 & 1.40\\
\addlinespace
ConditionalQA & \method{} (\jev{}) & W & 50 & 0.82/1.02 & 0.29\\
 & \quad + dense candidates & H & 50 & 0.85/1.13 & 0.29\\
 & Dense top-20 + GPT-5 & R & 285 & 14.4/31.4 & 1.20\\
\addlinespace
MTRAG & \method{} (\jev{}) & W & 24 & 1.76/2.26 & 1.73\\
 & \quad + dense candidates & H & 24 & 1.73/2.64 & 1.75\\
 & Dense top-20 + GPT-5 & R & 188 & 12.2/20.8 & 1.90\\
\addlinespace
Natural Questions & \method{} (\jev{}) & W & 100 & 1.73/2.54 & 2.33\\
 & \quad + dense candidates & H & 100 & 1.64/2.23 & 2.33\\
 & Dense top-20 + GPT-5 & R & 500 & 15.0/23.8 & 1.40\\
\bottomrule\end{tabular}\end{table}

\paragraph{Baseline timing.}
Locally served baselines (\cref{tab:local-latency}) are timed per query
after model and index loading and warm-up, with GPU synchronization, on
an RTX 3090 (BM25 on CPU). Dense top-100 + Qwen3-Reranker is timed on
one RTX A6000 on seeded samples of 48, 50, 24 and 100 queries
(MultiDoc2Dial, ConditionalQA, MTRAG, Natural Questions): query encoding
with Qwen3-Embedding-8B, exact top-100 search over the corpus vectors
(2.68M on Natural Questions), and reranking of the 100 candidates in
batches of up to 32,768 padded tokens. Reranking accounts for nearly all
of its latency; encoding and search take 0.05--0.08 seconds. GPT-5 (low
reasoning effort) reranks 20 saved dense candidates with up to 16 API
calls in flight; its timing and cost (rows R in \cref{tab:cost})
exclude dense retrieval and use a seeded sample of 1,000 validation
turns on MultiDoc2Dial and the full evaluation sets elsewhere. Because
hardware and samples differ across systems, latency ratios are not
controlled speedups.

\begin{table}[H]\centering\small
\caption{Median latency of the locally served baselines (seconds per
query, RTX 3090 unless noted; BM25 on CPU), including encoding and search. BM25 on
ConditionalQA indexes headings with the paragraphs; MTRAG uses the
six-turn input.}
\label{tab:local-latency}
\begin{tabular}{@{}lrrr@{}}\toprule
System & MultiDoc2Dial & ConditionalQA & MTRAG\\\midrule
BM25 & 0.001 & 0.001 & 0.008\\
SPLADE-v3 (DistilBERT) & -- & 0.117 & 0.298\\
SPLADE++ & -- & 0.090 & 0.177\\
Dense & 0.066 & 0.047 & 0.116\\
Dense top-100 + Qwen3-Reranker (A6000) & 7.51 & 3.33 & 14.33\\
\bottomrule\end{tabular}\end{table}

\paragraph{Offline preprocessing.}
Offline indexing is excluded from all online measurements. Encoding
the corpus with Qwen3-Embedding-8B took 180\,s for MultiDoc2Dial,
781\,s for ConditionalQA and 9,215\,s for MTRAG on an RTX 3090; SPLADE++
and SPLADE-v3 took 317\,s and 243\,s for MTRAG. \method{} encodes no
passages. Only on Natural Questions does building the hierarchy require
an offline pass: embedding each page's title and first sentence with
e5-base-v2 for clustering and describing the 1,000 clusters with Qwen3-8B
(\cref{app:datasets:nq}).

\section{Open-Weight Judge Training}
\label{sec:setup:train}

\paragraph{Model.}
The open-weight judges are Qwen3-Base models (1.7B, 4B and 8B) whose
language-model head is replaced by a scoring head. Each candidate is
posed as its own question with the answer options ``no'' and ``yes'',
appended to the shared context. All questions start at the position
right after the context and cannot attend to each other, so the
judgments of different candidates depend only on the shared context, as
in \cref{eq:binary}. The head projects the hidden states of the
question's end token and of each option's end token into a common
space, scales their cosine similarity by a learned inverse temperature
(capped at 100), and normalizes over the question's options with a
softmax; $P(\mathrm{yes})$ is the probability of ``yes''. Training
minimizes the cross-entropy over the options. In the general and
MultiDoc2Dial-only checkpoints, the end tokens of a question's two
options can also attend to each other; the LoRA checkpoint does not use
this link.

\paragraph{General training.}
The general models are trained on a mixture of public multiple-choice
and classification datasets, including reading comprehension (RACE,
QuALITY), science and commonsense question answering (SciQ, ARC,
CommonsenseQA, HellaSwag), natural language inference (MNLI), intent
and topic classification, and dialogue, tool-use and web-navigation
decisions (MultiWOZ, xLAM, Mind2Web). The label of each example is the
dataset's gold option; the alternatives are its own distractors or, for
classification, a random subset of the class labels. The mixture
contains no MultiDoc2Dial or ConditionalQA data.

\paragraph{Retrieval training data.}
MultiDoc2Dial training examples are generated from its training split
with the routing, page and passage stages of the navigation; 3\% of
training dialogues are held out for monitoring. The generator simulates
the notepad with a gold page with probability 0.65, includes the gold
page among three candidates with probability 0.85, and adds a second
help-center negative stage with probability 0.17; identical-text copies
of the gold passage are never used as negatives. ConditionalQA examples
are built by the navigation's own page and paragraph stages, so their
inputs match those at inference. A page-stage example lists all 652
pages with ``yes'' for the gold page; a paragraph-stage example lists
every paragraph of five candidate pages with ``yes'' for the evidence
paragraphs. The gold page's rank among the five candidates is drawn
from \jev{}'s page-stage ranks on the 300 ConditionalQA design
questions, and the remaining candidates are the top BM25 pages; when
the gold page is not drawn, all paragraphs are negatives. Paragraphs
whose exact text appears as evidence on another page are not asked, and
questions without evidence contribute only a page-stage example.
Questions on 10\% of the training pages are held out for monitoring.

\paragraph{Optimization.}
The fine-tuning runs on retrieval data use AdamW ($\beta_1=0.9$,
$\beta_2=0.95$, no weight decay), gradient clipping at 1.0, 50 warm-up
steps and cosine decay to 10\% of the peak learning rate. The \emph{MD2D-only} model trains the head
alone for 200 steps (learning rate $10^{-3}$) and then the full model
for one epoch; \emph{general $\to$ MD2D} continues from the general 8B
checkpoint for one epoch without the head-only phase. Both use learning rates of $10^{-5}$ (backbone) and $10^{-4}$ (head), a
maximum context of 32,768 tokens and packed rows of 40,960 tokens;
overlong passage-stage records (5.5\%) are dropped rather than
truncated. The \method{} (8B trained) model reported in
\cref{tab:trained-judge} instead continues the general 8B model with
LoRA (rank 64, $\alpha=128$, on the attention and MLP projections of
every layer) and a fully trained head, both at learning rate
$10^{-4}$, for one epoch over 30\% of the MultiDoc2Dial training
dialogues (a seeded sample) and all ConditionalQA training examples,
with rows of 65,536 tokens.

\paragraph{Results.}
The LoRA model is the only checkpoint evaluated on both datasets; the
checkpoints trained on MultiDoc2Dial alone reach higher MultiDoc2Dial
recall (\cref{tab:judge-full}). All models are evaluated with the
retrieval configurations of \cref{app:retrieval}, ranking by their
binary judgments.

\begin{table}[H]\centering\small
\caption{Open-weight judges (\%). \emph{General}: trained without
MultiDoc2Dial or ConditionalQA data; \emph{+ MD2D + CQA (LoRA)}: the
\method{} (8B trained) model of \cref{tab:trained-judge}.}\label{tab:judge-full}
\begin{tabular}{@{}llrrr@{}}\toprule
\multicolumn{5}{@{}l}{\textit{MultiDoc2Dial validation}}\\
Judge / training & Pages & R@1 & R@5 & R@10\\\midrule
1.7B / general & 3 & 20.5 & 38.3 & 44.2 \\
1.7B / general & 10 & 12.8 & 28.7 & 36.3 \\
4B / general & 3 & 28.5 & 52.9 & 59.8 \\
4B / general & 10 & 21.9 & 47.2 & 56.5 \\
8B / general & 3 & 40.4 & 66.0 & 71.9 \\
8B / general & 10 & 37.3 & 65.3 & 73.0 \\
8B / MD2D only & 3 & 63.1 & 84.2 & 86.6 \\
8B / MD2D only & 10 & 64.3 & 87.6 & 91.3 \\
8B / general $\to$ MD2D & 3 & 63.5 & 84.1 & 86.4 \\
8B / general $\to$ MD2D & 10 & 65.8 & 87.9 & 91.2 \\
8B / general + MD2D + CQA (LoRA) & 3 & 60.7 & 82.7 & 85.8 \\
8B / general + MD2D + CQA (LoRA) & 10 & 60.8 & 86.1 & 90.0 \\
\midrule
\multicolumn{5}{@{}l}{\textit{ConditionalQA development (271 questions with evidence)}}\\
Judge / training & Pages & Hit@1 & R@5 & R@10\\\midrule
8B / general & 3 & 35.4 & 36.6 & 44.5 \\
8B / general & 5 & 38.4 & 37.3 & 45.5 \\
8B / general + MD2D + CQA (LoRA) & 3 & 75.3 & 66.7 & 78.8 \\
8B / general + MD2D + CQA (LoRA) & 5 & 77.1 & 67.7 & 80.2 \\
\bottomrule\end{tabular}\end{table}

\end{document}